\documentclass[11pt]{article}

\usepackage[margin=1in]{geometry}

\usepackage[T1]{fontenc}
\usepackage[utf8]{inputenc}
\usepackage[french]{babel}

\usepackage{amsmath}
\usepackage{amssymb}

\usepackage{graphicx}

\usepackage[ruled,vlined,linesnumbered]{algorithm2e}

\usepackage{tikz}
\usetikzlibrary{arrows.meta, positioning, shapes.geometric, calc, fit}

\usepackage{xcolor}

\usepackage{hyperref}

\usepackage{placeins}

\usepackage{enumitem}

\usepackage{float}

\title{
    \textbf{Curriculum-Based Reinforcement Learning for Autonomous UAV Navigation in Unknown Curved Tubular Conduits}
}

\author{
    Zamirddine Mari$^{1}$,
    Jérôme Pasquet $^{2}$,
    Julien Seinturier$^{3}$ \\
    \\
    \small $^{1}$DGA Techniques Navales - Direction Générale de l'Armement, Toulon, France \\
    \small $^{2}$LIRMM, TETIS, CNRS, Université de Montpellier Paul-Valéry, Montpellier, France \\
    \small $^{3}$LIS, CNRS, Université de Toulon, Toulon, France \\
    \small Corresponding author: zamirddine.mari@intradef.gouv.fr
}

\date{} % pas de date, style arXiv

\begin{document}

\maketitle

% --- Mention préprint ---
\begin{center}
\textit{This document is a preprint prepared for submission to Sensors (MDPI).\\
The title and content are preliminary and may be updated in future versions.}
\end{center}
\bigskip

\renewcommand{\abstractname}{Abstract}
\begin{abstract}
Autonomous drone navigation in confined tubular environments remains a major challenge due to the constraining geometry of the conduits, the proximity of the walls, and the perceptual limitations inherent to such scenarios. We propose a reinforcement learning approach enabling a drone to navigate unknown three-dimensional tubes without any prior knowledge of their geometry, relying solely on local observations from LiDAR and a conditional visual detection of the tube center. In contrast, the \textit{Pure Pursuit} algorithm, used as a deterministic baseline, benefits from explicit access to the \textit{centerline}, creating an information asymmetry designed to assess the ability of RL to compensate for the absence of a geometric model.

The agent is trained through a progressive Curriculum Learning strategy that gradually exposes it to increasingly curved geometries, where the tube center frequently disappears from the visual field. A turning-negotiation mechanism, based on the combination of direct visibility, directional memory, and LiDAR symmetry cues, proves essential for ensuring stable navigation under such partial observability conditions.

Experiments show that the PPO policy acquires robust and generalizable behavior, consistently outperforming the deterministic controller despite its limited access to geometric information. Validation in a high-fidelity 3D environment further confirms the transferability of the learned behavior to a continuous physical dynamics.

The proposed approach thus provides a complete framework for autonomous navigation in unknown tubular environments and opens perspectives for industrial, underground, or medical applications where progressing through narrow and weakly perceptive conduits represents a central challenge.
\end{abstract}

% --- Mots-clés (option arXiv) ---
\bigskip

%
%
% --- INTRODUCTION ---
\section{Introduction}

Autonomous drone navigation in confined and tubular environments has become a major challenge for many applications, such as infrastructure inspection, exploration of technical ducts, analysis of underground networks, or search-and-rescue missions. Recent studies have highlighted the growing interest in autonomous drone flight within tunnels or narrow structures, emphasizing the intrinsic complexity of such highly constrained environments~\cite{ref_1, ref_2}. Other works have shown the difficulties associated with underground or poorly lit environments, where onboard perception is essential to ensure safe progression~\cite{ref_3, ref_4}, while approaches based on tilted LiDARs have been proposed to improve navigation in tunnels with complex geometries~\cite{ref_5}.

Navigating through an unknown tube \emph{a priori} requires overcoming several major challenges: absence of GPS, aerodynamic turbulences caused by the immediate proximity of the walls, abrupt variations in curvature, as well as severe perceptual limitations due to darkness or lack of texture. These characteristics bring this problem close to endoscopic navigation, where robotic systems must evolve within narrow, weakly textured, and potentially tortuous conduits. In this field, several studies have demonstrated the relevance of reinforcement learning for achieving safe and adaptive progression in complex geometries~\cite{ref_6, ref_7, ref_8}.

In this study, the \textit{Pure Pursuit} algorithm is used as a \emph{deterministic baseline}. This algorithm, widely employed in mobile and aerial robotics, provides robust trajectory tracking when a reference path is available~\cite{ref_9}. In our experimental setting, Pure Pursuit benefits from privileged access to the tube's \emph{centerline}, i.e., the 3D curve used to generate the conduit geometry. In contrast, the reinforcement learning agent has no prior knowledge of the tube geometry and must learn to navigate exclusively from its local observations. This deliberate asymmetry allows for a rigorous evaluation of RL’s ability to ensure safe navigation in unknown environments potentially more complex than those anticipated by deterministic methods.

Beyond learning a global navigation policy, a key aspect of our approach lies in the handling of turns, which represent the most critical situations in a complex tubular environment. When the tube center disappears from the visual field during directional changes, the agent must rely on a subtle combination of sensory information (LiDAR, camera, memory of the last known direction) to maintain both centering and alignment. This turning-negotiation problem strongly structures the modeling proposed in this article.

In this context, the main contributions of this work are as follows:
\begin{itemize}
    \item (1) the development of a reinforcement learning agent capable of navigating in unknown tubular environments without any prior knowledge of the tube geometry;
    \item (2) the design of a three-dimensional simulation environment that can generate tubes of varying complexity from synthetic guiding curves;
    \item (3) an in-depth experimental comparison between the RL agent and a Pure Pursuit method directly exploiting the tube centerline.
\end{itemize}

Section~\ref{sec:methodology} details all modeling choices made to make this problem learnable, from the definition of the action space and the turning-negotiation mechanism to the construction of the observation space and the formulation of the reward function.

\section{Méthodologie}
\label{sec:methodology}

\subsection{Problem Description}

The problem addressed consists in enabling an autonomous drone to navigate within a three-dimensional tubular
environment whose geometry may exhibit significant variations in curvature. The drone must traverse the tube
until its endpoint while avoiding any collision with the internal walls and maintaining a stable trajectory,
despite having no global information about the shape of the conduit.

The tube is generated from a smooth spatial curve acting as a guiding axis. This curve may present substantial
local variations: abrupt changes in orientation, tightly curved regions, or portions temporarily unobservable
from the drone’s current position. Turns therefore constitute the most critical situations, as they can cause a
temporary loss of visibility of the central point of the conduit within the drone’s field of view.

The drone evolves under realistic kinematics: it possesses a forward direction, a controllable speed, and a
continuously updated local frame. No map of the tube nor any prior knowledge of its geometry is provided. Its
perception relies exclusively on local observations, including:
\begin{itemize}
    \item proprioceptive information (orientation, forward direction, velocity, estimated progression);
    \item limited exteroceptive measurements, in particular a front/back perimeter LiDAR providing normalized
          distances to the walls;
    \item a basic visual module detecting, when possible, a point corresponding to the local center of the
          conduit in the field of view.
\end{itemize}

These elements constitute the perceptual basis enabling navigation, centering, and adaptation to geometric
variations until the drone reaches the end of the tube.

\subsection{General Description of the Approach}

This subsection presents the principles guiding our navigation strategy before introducing a mathematical 
formalization within the framework of a Markov Decision Process.

At each time step, the agent constructs a perceptual state synthesizing all available local observations. This 
state includes:
(i) instantaneous kinematics (orientation, forward direction, speed),
(ii) the current progression along the tube,
(iii) the possible detection of a target point visible in the field of view,
(iv) and features derived from the front/back LiDAR, enabling inference of the local symmetry of the conduit 
and anticipation of upcoming curved regions.

When the target point is visible, the drone learns to orient itself preferentially toward this reference 
direction. When the target temporarily disappears during a turn, a short-term directional memory preserves the 
last useful orientation, ensuring a smooth transition until the structure becomes observable again. In the 
absence of reliable visual or memory cues, the tube geometry is estimated from LiDAR asymmetries.

The drone's motion is thus governed by the joint adjustment of its speed and forward direction based on local 
observations and relevant geometric cues. Behaviors are evaluated according to instantaneous criteria of 
centering, alignment, and consistency with the implicit structure of the conduit, as well as the ability to 
negotiate turns effectively.

Finally, to provide a deterministic point of comparison, the agent’s performance is contrasted with that of a 
trajectory-following algorithm of the \emph{Pure Pursuit} type, which directly exploits the centerline used to 
generate the tube.

The next section formalizes this decision framework using a \emph{Markov Decision Process}.

\subsection{Problem Formulation}

The autonomous navigation problem of a drone inside a confined tubular environment is formulated as a \emph{Markov Decision Process} (MDP), defined by the tuple
$\mathcal{M} = \langle \mathcal{S}, \mathcal{A}, \mathcal{P}, r, \gamma \rangle$.
Here, $\mathcal{S}$ denotes the state space, $\mathcal{A}$ the action space, $\mathcal{P}(s'|s,a)$ the stochastic transition dynamics,
$r(s,a)$ the reward function, and $\gamma \in (0,1]$ the discount factor.
The objective of the agent is to maximize the expected cumulative reward:
\begin{equation}
    J(\pi) = \mathbb{E}_{\pi}\left[ \sum_{t=0}^{\infty} \gamma^t r(s_t, a_t) \right],
\end{equation}
where $\pi(a|s)$ denotes the agent’s parameterized policy.

\subsubsection{PPO Algorithm for Reinforcement Learning}
Among policy optimization methods, the \emph{Proximal Policy Optimization} (PPO) algorithm has become a widely adopted reference due to its stability and performance~\cite{ref_10}.

\medskip
\noindent
The MDP formulation provides a general mathematical framework for learning through interaction. To clarify how this formalism applies to navigation in tubular environments, we now describe the physical components of the system: the drone, its sensors, and the local structure of the environment, which jointly determine both the state space and the dynamics of the problem.

\subsection{Drone Modeling}
\begin{figure}[h!]
    \centering
    \includegraphics[width=0.5\linewidth]{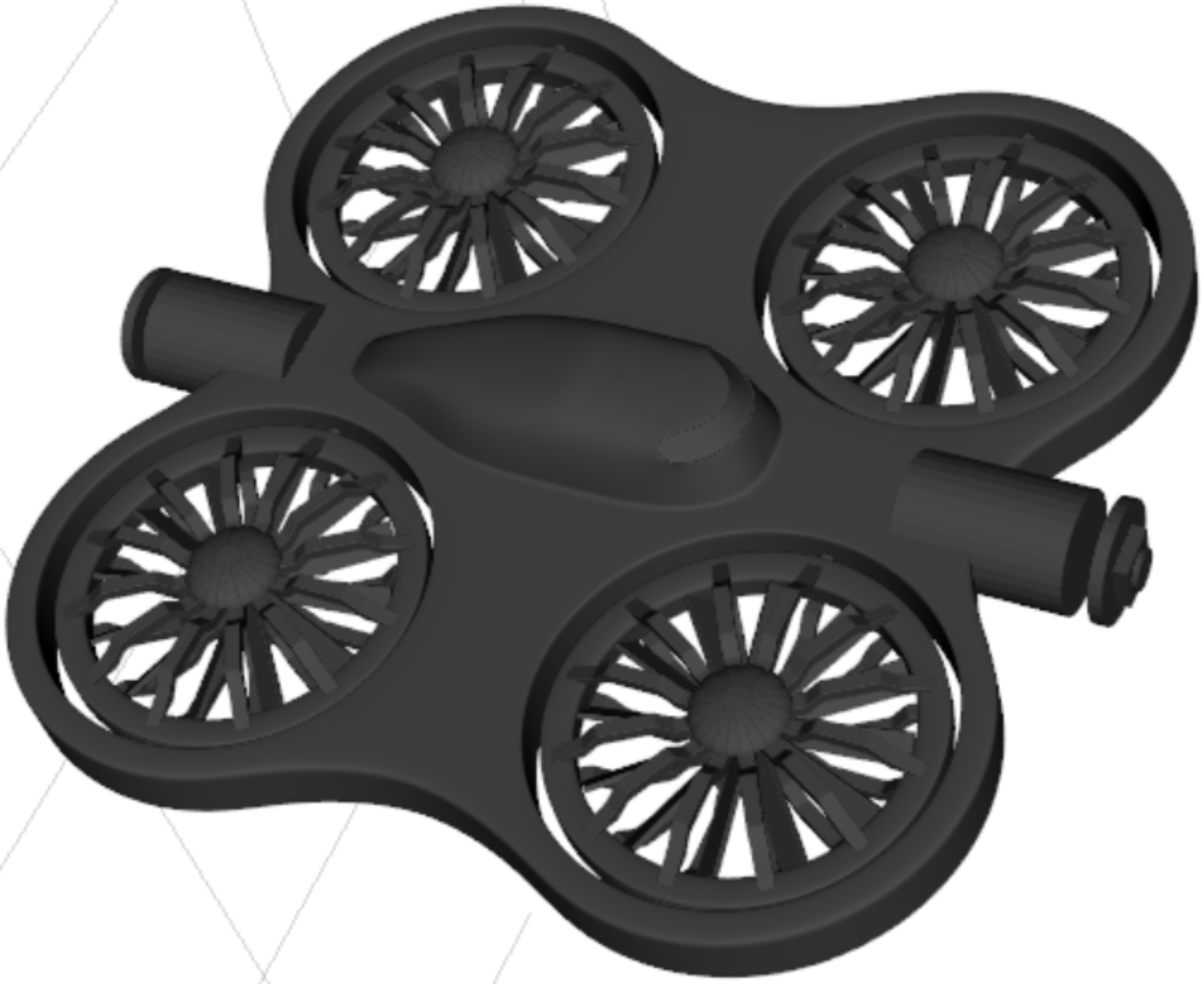}
    \caption{Illustration of the 3D model of the drone used in the study.}
    \label{fig:drone3D}
\end{figure}

\begin{figure}[h!]
    \centering
    \includegraphics[width=1.0\linewidth]{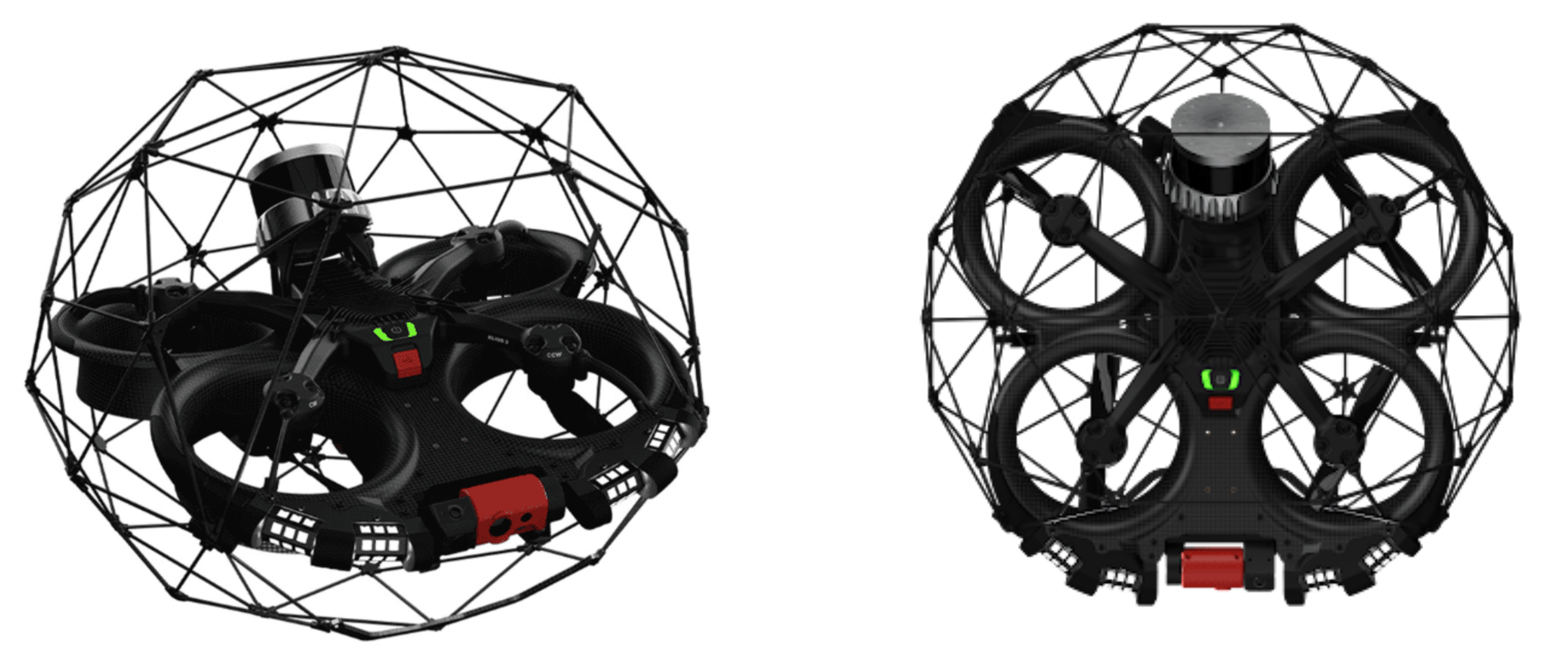}
    \caption{Photographs of the Flyability ELIOS 3 quadcopter equipped with a front-facing camera and a rear-tilted Ouster OS0 LiDAR.}
    \label{fig:elios3}
\end{figure}

In this work, as illustrated in Figure~\ref{fig:drone3D}, the drone is modeled as an autonomous quadcopter required to navigate through a narrow tube, negotiate turns, and avoid collisions. The modeling choice is inspired by the sensory configuration of commercial drones specialized in confined-structure inspection, such as the Flyability ELIOS 3 shown in Figure~\ref{fig:elios3}, designed for the exploration of tunnels, caves, and constrained industrial environments.

\begin{figure}[h!]
    \centering
    \includegraphics[width=0.7\linewidth]{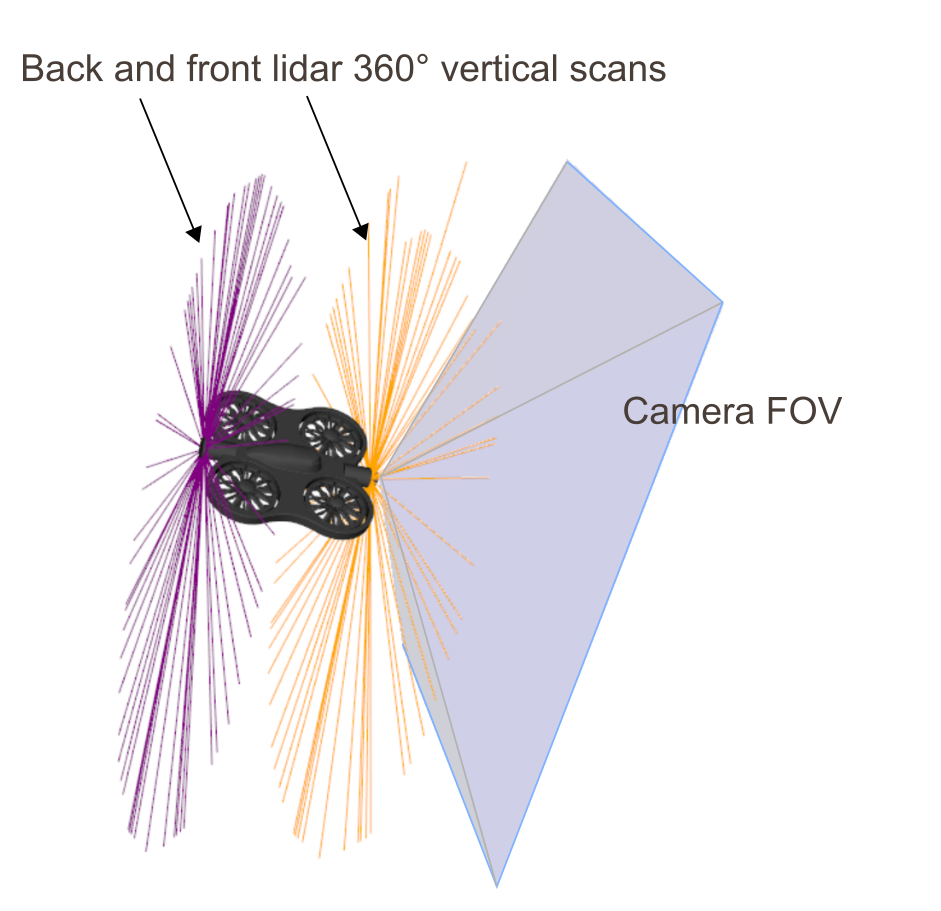}
    \caption{Illustration of the drone equipped with its front-facing camera and its front and rear LiDARs.}
    \label{fig:drone_capt}
\end{figure}

As shown in Figure~\ref{fig:drone_capt}, the theoretical drone considered in this study is equipped with the following perception sensors:
\begin{itemize}
    \item a front-facing camera, aligned with the longitudinal axis of the quadcopter and used to acquire images of the tube and detect its center when visible;
    \item two LiDARs, positioned at the front and rear of the drone, each providing $360^\circ$ coverage on a vertical plane perpendicular to the drone’s longitudinal axis, offering robust perception of radial distances to the tube walls.
\end{itemize}

This sensor configuration is consistent with recent practices in autonomous drone navigation within tubular environments. For instance, in~\cite{ref_12}, the authors propose a navigation approach based on a front-facing camera detecting the tunnel center in each image and controlling the drone’s heading accordingly. Complementarily, several works~\cite{ref_13, ref_14} leverage LiDAR configurations to build a local geometric representation of the environment and provide reliable obstacle-distance estimation in underground environments.

In the present paper, we assume that the drone’s front camera can detect the tunnel center when it lies within its field of view, an assumption directly inspired by the approach in~\cite{ref_12}. The LiDARs complement this perception by providing continuous measurements of distances to the walls over the full vertical perimeter, which is essential for assessing navigability and correcting trajectory deviations when the tunnel center is not observable.

This camera–LiDAR combination, inspired both by industrial solutions (ELIOS 3) and recent scientific practices, offers a robust compromise between visual perception and geometric measurement for autonomous navigation in confined tubular environments.

\vspace{0.2cm}

\subsection{Environment}

\begin{figure}[h!]
    \centering
    \includegraphics[width=0.7\linewidth]{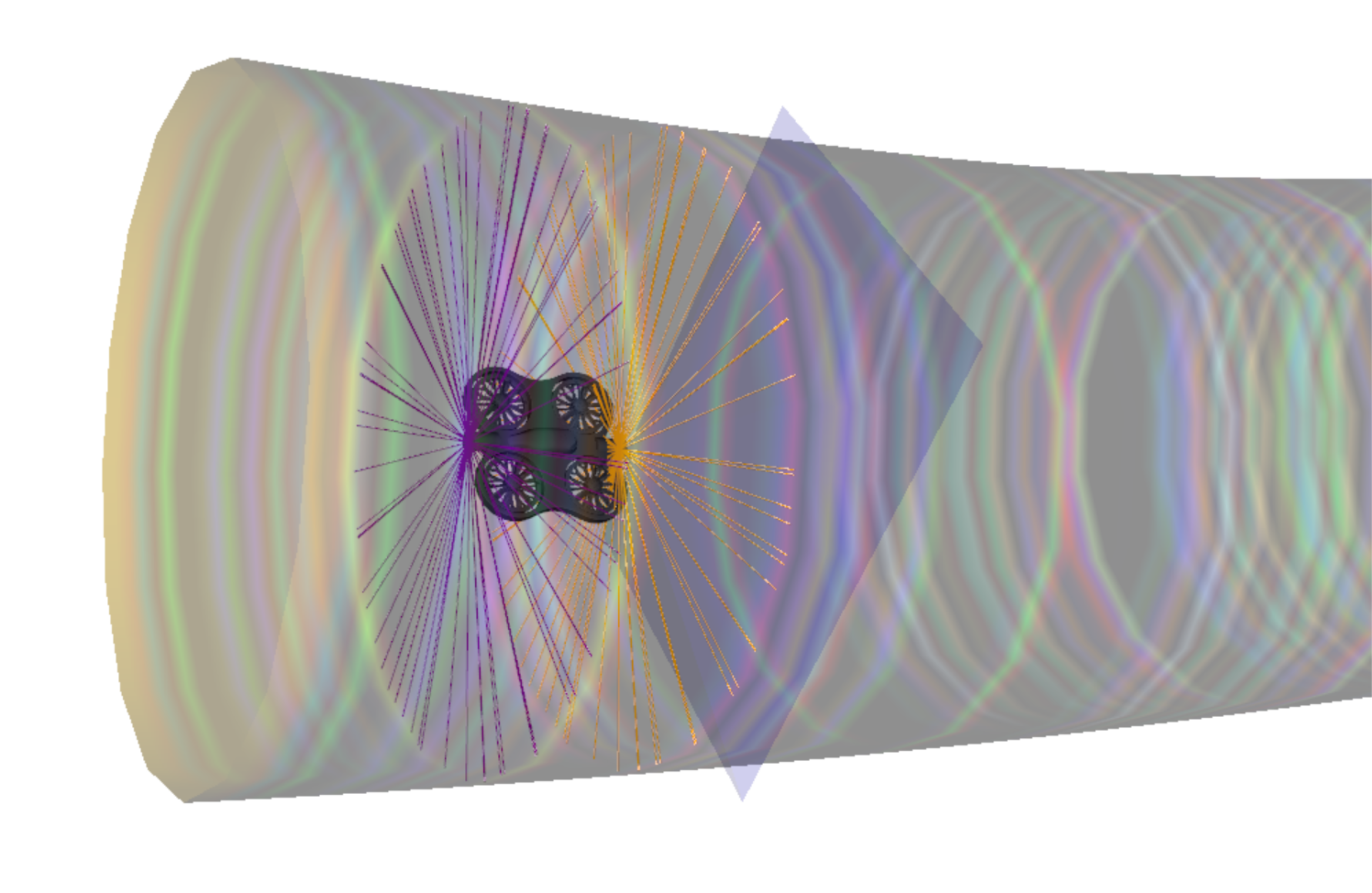}
    \caption{Illustration of the drone's evolution within the tubular environment.}
    \label{fig:env_tube}
\end{figure}

As illustrated in Figure~\ref{fig:env_tube}, the environment is modeled as a confined space $\mathcal{E} \subset \mathbb{R}^3$ in which the drone evolves. The kinematic state of the quadcopter at time $t$ is represented by the vector:

\begin{equation}
    s_t = \begin{bmatrix} \mathbf{p}_t \\ v_t \\ \mathbf{R}_t \end{bmatrix} \in \mathcal{S},
\end{equation}

where:
\begin{itemize}
    \item $\mathbf{p}_t = [x_t, y_t, z_t]^\top \in \mathbb{R}^3$ is the position of the center of mass in the global reference frame,
    \item $v_t \in \mathbb{R}$ is the drone’s speed along its longitudinal (forward) axis,
    \item $\mathbf{R}_t \in SO(3)$ is the rotation matrix representing the drone’s orientation with respect to the global frame, forming a local orthonormal basis.
\end{itemize}

This representation captures the drone’s spatial position, forward speed, and orientation—elements essential for trajectory planning and autonomous navigation control in complex tubular environments.

\medskip
\noindent \textbf{Curriculum Learning for progressively increasing tube complexity}

To facilitate reinforcement learning and improve policy robustness, we introduce a \emph{Curriculum Learning} (CL) mechanism.  
The central idea is to progressively expose the agent to increasingly complex tubular environments while controlling the geometric difficulty of the curves used to generate the 3D conduits.

\begin{figure}[h!]
    \centering
    \includegraphics[width=1.0\linewidth]{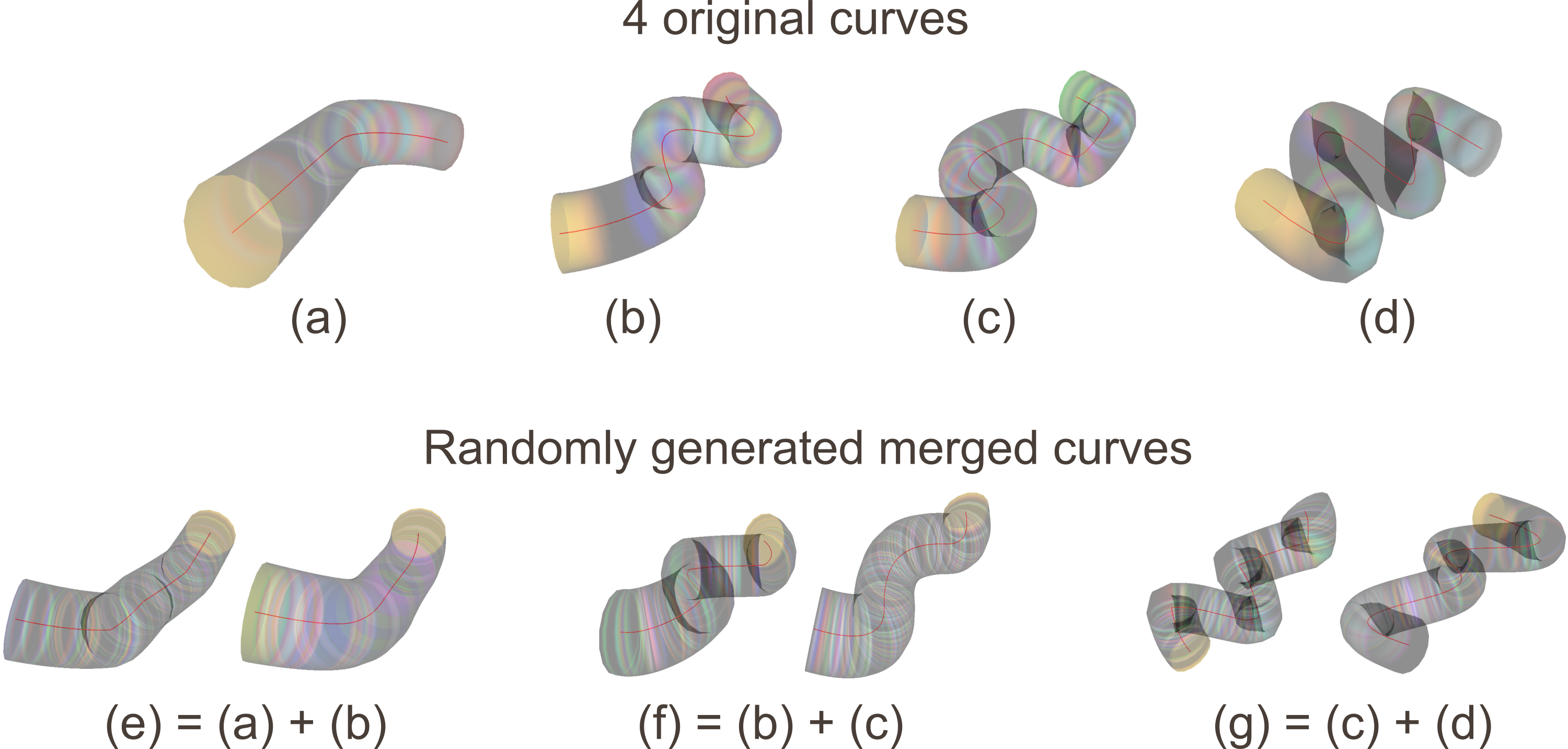}
    \caption{Illustration of the massive tube-generation principle during training.}
    \label{fig:mass_gen}
\end{figure}

As shown in Figure~\ref{fig:mass_gen}, tube generation relies on a set of raw curves derived from predefined synthetic models.  
At each episode, two curves are selected according to the curriculum level, smoothed via spline interpolation, and combined to produce a mixed tubular geometry.  
This process follows two main steps:

\begin{itemize}
    \item \textbf{uniformization and smoothing}: each curve is reparameterized via B-splines with random tension and degree, ensuring geometric diversity even within the same difficulty level;
    \item \textbf{curriculum-controlled selection}:
    \begin{itemize}
        \item level 0: nearly straight curves, indices $\{(a),(b)\}$;
        \item level 1: moderately curved shapes, indices $\{(b),(c)\}$;
        \item level 2: highly curved shapes, indices $\{(c),(d)\}$.
    \end{itemize}
    \item \textbf{geometric fusion}: two curves are interpolated using a random factor $\alpha \in [0.1, 0.9]$ to produce a new trajectory at each episode, even within the same curriculum level.
\end{itemize}

This procedure ensures a progressive, controlled, yet non-repetitive increase in geometric complexity.  
It notably exposes the agent to difficult scenarios involving low-visibility turns, where the vanishing point temporarily disappears and navigation must rely solely on inertial and LiDAR cues.

The agent moves to the next level only when the average success rate exceeds a predefined threshold. Table~\ref{tab:curriculum_tubes} summarizes the levels used.

\begin{table}[h!]
\centering
\caption{Curriculum Learning for navigation in curved tubes: complexity and success thresholds}
\label{tab:curriculum_tubes}
\begin{tabular}{c c c}
\hline
\textbf{Level} & \textbf{Complexity (index)} & \textbf{Success threshold} $\eta_{\text{succ}}$ \\
\hline
0 & 0 & 0.85 \\
1 & 1 & 0.80 \\
2 & 2 & 0.80 \\
\hline
\end{tabular}
\end{table}

\medskip
\noindent\textbf{Justification of the thresholds.}  
The success thresholds used in this curriculum were deliberately chosen to be \emph{moderate}. In this initial development phase, the primary objective is not to maximize the controller’s absolute performance but rather to ensure stable RL policy convergence while validating the feasibility of the approach across increasingly complex environments.

Stricter thresholds (e.g., $>0.9$) would increase the risk of stagnation in intermediate levels, particularly when the vanishing point is lost in strongly curved tubes. They would also significantly increase the training time required to progress between levels, making the curriculum less practical for this preliminary study.

The adopted thresholds (0.80–0.85) therefore represent an effective compromise: high enough to enforce meaningful progressive learning, yet accessible enough to allow smooth exploration and full validation of the RL approach.

\medskip
In summary, Curriculum Learning serves as a key mechanism to:
\begin{enumerate}
    \item stabilize RL policy convergence by avoiding excessively complex scenarios at the start,
    \item improve the drone’s ability to generalize to tubes with varied shapes and tight turns,
    \item provide a progressive training framework suited to the interaction between the nominal Pure Pursuit controller and the adaptive RL module.
\end{enumerate}

\subsection{Observation Space}

At each time step, the agent receives an observation vector
\[
\mathbf{o}_t \in [-1,1]^{37},
\]
constructed from five components: features derived from the LiDAR scans, the drone’s orientation and kinematics, the visual perception of the target, a memory mechanism for handling turning phases, and a set of global context indicators.

\subsubsection{Derived LiDAR Features}

Rather than relying directly on raw LiDAR measurements,  
the environment uses a set of nine geometric features extracted from the front and rear scans:
\[
\mathbf{r}_t =
\big[
h_f, v_f,
h_r, v_r,
s_f, s_r,
m_f, m_r,
\ell_{\min}
\big] \in [-1,1]^9.
\]
They include:
\begin{itemize}
    \item horizontal and vertical asymmetries for the front $(h_f, v_f)$ and rear $(h_r, v_r)$ scans;
    \item front and rear symmetry scores $s_f$ and $s_r$;
    \item average distances $m_f$ and $m_r$;
    \item the normalized minimum distance $\ell_{\min}$, used as a safety indicator.
\end{itemize}
These quantities summarize the essential LiDAR information while removing the local variability of raw measurements.

\subsubsection{Orientation and Kinematics}

The drone provides its local orthonormal basis
\[
(\mathbf{f}_t, \mathbf{u}_t, \mathbf{r}_t) \in [-1,1]^9,
\]
corresponding respectively to its forward, upward, and lateral axes.  
In addition, the observation includes:
\begin{itemize}
    \item the normalized longitudinal velocity;
    \item its increment $\Delta v_t$ between two time steps;
    \item the global progression along the tube (between $-1$ and $1$);
    \item the drone’s position normalized by the tube radius.
\end{itemize}
This subset contributes 15 dimensions.

\subsubsection{Visual Perception of the Target}

If the tube center is visible in the field of view, its direction is projected into the drone’s local frame:
\[
\mathbf{d}^{\mathrm{cam}}_t \in [-1,1]^3,
\]
along with:
\begin{itemize}
    \item a normalized depth of the target;
    \item a binary visibility flag ($+1$ if visible, $-1$ otherwise).
\end{itemize}
This module provides 5 dimensions.

\subsubsection{Memory Mechanism}

When a turn temporarily hides the target, the agent relies on:
\begin{itemize}
    \item a temporal ratio indicating how long the target has been invisible;
    \item the last memorized target direction, expressed in the local frame (3 values);
    \item a memory-availability flag ($+1$ or $-1$).
\end{itemize}
This block adds another 5 dimensions.

\subsubsection{Global Context}

Finally, three additional indicators complete the state:
\begin{itemize}
    \item a secondary safety metric based on the LiDAR minimum distance;
    \item an indicator of whether the drone remains inside the tube;
    \item the normalized progression within the episode.
\end{itemize}

\subsubsection{Complete Vector}

The final observation is therefore:
\[
\mathbf{o}_t =
\big[
\underbrace{\mathbf{r}_t}_{9},
\underbrace{\text{kinematics}}_{15},
\underbrace{\text{camera}}_{5},
\underbrace{\text{memory}}_{5},
\underbrace{\text{context}}_{3}
\big]
\in [-1,1]^{37},
\]
a compact state composed of 37 informative features specifically designed for navigation in confined tubular environments.

\subsection{Action Space}

\begin{figure}[h!]
    \centering
    \includegraphics[width=1.0\linewidth]{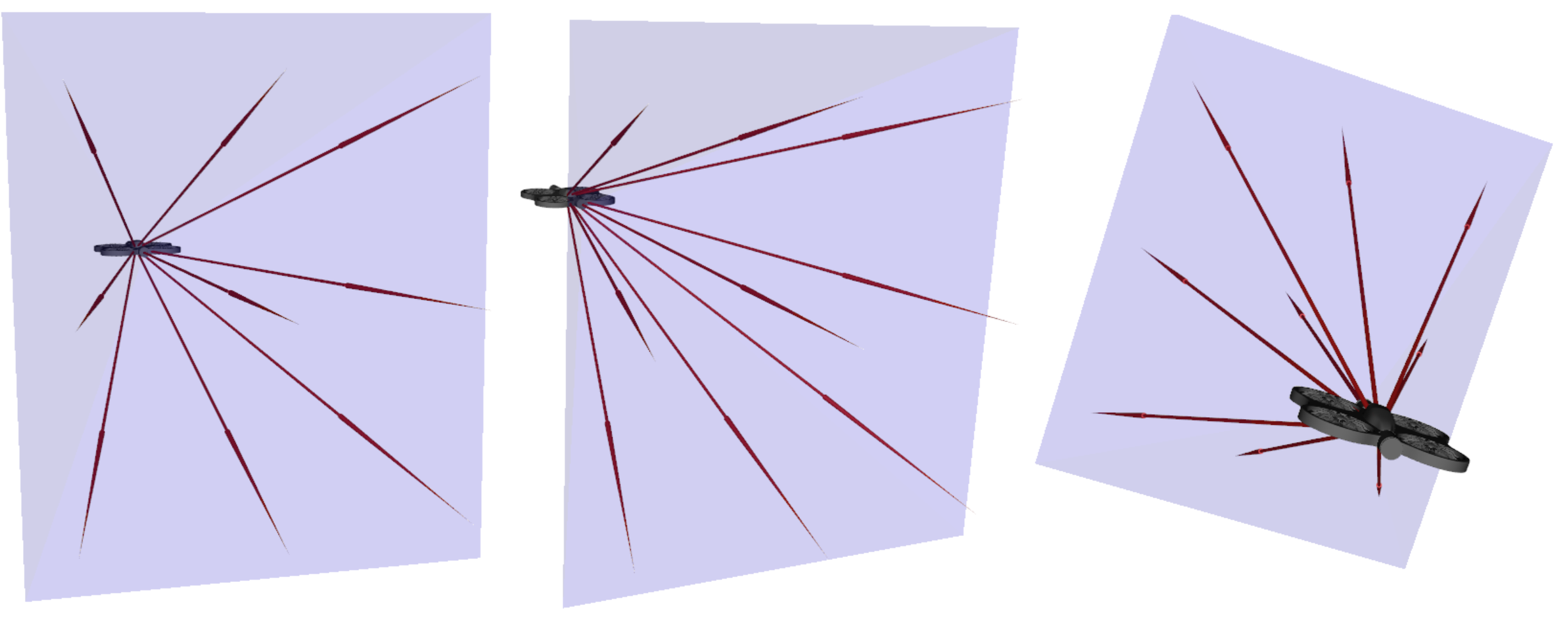}
    \caption{Uniform distribution of action directions within the camera’s field-of-view cone.}
    \label{fig:action_space}
\end{figure}

In the context of autonomous navigation in confined tubular environments, the agent must continuously move forward while 
applying corrective orientation adjustments to follow the conduit’s geometry. The action space, illustrated in 
Figure~\ref{fig:action_space}, was designed to satisfy this constraint: available orientations are restricted to a cone 
consistent with the front camera’s field of view (FOV), ensuring that each action directs the drone toward plausible 
continuations of the conduit—even in phases where the tube center is no longer visible.  
This design allows the agent to explore controlled, geometrically plausible directions.

At each step, the action corresponds to an \emph{orientation–speed} pair chosen from a discrete space:
\[
\mathcal{A} = \mathcal{D} \times \mathcal{V},
\qquad |\mathcal{A}| = 9 \times 4 = 36,
\]
where $\mathcal{V}$ contains four levels of longitudinal speed, and $\mathcal{D}$ contains nine orientations uniformly 
distributed within the angular limits of the FOV.

\paragraph{Definition of the orientation cone.}
The maximum half-angles of the cone are given by:
\[
\theta_{\max}^{(a)} = \arctan\!\left(\frac{f_w}{c}\right),
\qquad
\theta_{\max}^{(b)} = \arctan\!\left(\frac{f_h}{c}\right),
\]
where $f_w$, $f_h$, and $c$ denote respectively the half-width, half-height, and synthetic focal length of the camera.

\paragraph{Uniform distribution of directions.}
The set $\mathcal{D}$ contains nine angle pairs $(\alpha,\beta)$ defined as:
\[
(\alpha,\beta) \in 
\Big\{
(0,0),
\ \pm k\,\theta_{\max}^{(a)},
\ \pm k\,\theta_{\max}^{(b)},
\ (\pm k\,\theta_{\max}^{(a)},\pm k\,\theta_{\max}^{(b)})
\Big\},
\]
where $k=0.75$ ensures sufficiently strong corrections while preserving flight stability.
These orientations are symmetrically and uniformly distributed inside the cone, providing a reduced yet expressive 
set of corrective directions consistent with the FOV geometry.

\paragraph{Construction of the action direction.}
Each pair $(\alpha,\beta)$ is converted into a unit orientation vector by combining the drone’s current forward 
direction with two locally defined transverse axes.  
An additional constraint prevents the selection of backward-facing orientations, which would be incompatible with 
forward navigation inside a narrow conduit.

\paragraph{Justification of the design choice.}
This discrete action space is not intended to uniformly sample all possible 3D orientations. Rather, it provides a 
symmetric, uniformly distributed, and limited set of plausible directions for following the conduit.  
It allows the agent to:
\begin{itemize}
    \item explore orientations consistent with the tube’s geometric continuity, even when the center is temporarily invisible;
    \item apply sufficient corrections to follow curved sections;
    \item keep the action space compact, which favors efficient learning.
\end{itemize}

The next section describes how the agent exploits this action space to handle straight segments, anticipate turns, and 
maintain a stable trajectory even when the tube center becomes temporarily unobservable.
\subsection{Turn Negotiation}

As previously introduced, negotiating turns within a curved conduit is a major challenge for an agent equipped with a single forward-facing camera, whose field of view cannot guarantee continuous visibility of the geometric center of the tube. The strategy developed in this work relies on a progressive interplay between visual perception, directional memory, and geometric analysis from LiDAR measurements, allowing continuous adaptation to the local dynamics of the conduit.

\paragraph*{1) Early Turn Detection via Degradation of Visual Alignment}

Even before the target leaves the field of view, the tube curvature induces a lateral drift of the target point in the image plane. This drift is reflected in practice as a progressive decrease of the visual alignment score, computed as the dot product between the drone’s instantaneous forward direction and the vector from the camera to the target:
\begin{equation}
A_{\mathrm{vis}} = \left\langle \widehat{\mathbf{d}}_{\mathrm{drone}},\, \widehat{\mathbf{d}}_{\mathrm{cible}} \right\rangle .
\end{equation}

This indicator provides an explicit signal as long as the target remains visible. A significant reduction in $A_{\mathrm{vis}}$ reveals a local geometry unfavorable to frontal alignment, indicating that a turn is imminent.

This initiates a fully operational transitional phase: the drone actively adjusts its orientation in response to the degradation of visual centering, beginning to negotiate the turn before the target completely disappears from view. This phase is essential to reduce oscillations and to initiate a smooth rotation of the camera toward the new direction of the tube.

\paragraph*{2) Memory–Vision Transition When the Target Leaves the Field of View}

When curvature becomes sufficient to push the target outside the camera field of view, the system retains the last valid direction:
\begin{equation}
\widehat{\mathbf{d}}_{\mathrm{ref}} = \widehat{\mathbf{d}}_{\mathrm{mem}}.
\end{equation}

This directional memory, maintained as long as it remains recent, provides the dynamic continuity needed to enter the turn smoothly, extending the strategy initiated during the visual transitional phase.

\paragraph*{3) Geometric Navigation in Prolonged Absence of Vision}

If the memory expires, navigation relies solely on the instantaneous analysis of front and rear LiDAR measurements. In a turn, a natural mechanical dissymmetry emerges: the drone’s front tends to remain oriented toward the previously perceived or memorized direction, while the rear, subject to rotational motion and slight lateral drift, tends to move toward the outer wall of the curve.

This dynamic makes the front LiDAR measurements less reliable for assessing lateral safety, as they partly reflect the previously followed direction and become ambiguous in tight curvature. In contrast, the rear LiDAR provides more relevant geometric information: it indicates the drone’s true lateral offset from the walls during rotation, acting similarly to a tactile perception.

Thus, in the absence of vision, the front primarily contributes to orientation, while the rear plays a stabilizing role by preventing collisions with the tube walls during the maneuver. This functional dissymmetry enables the drone to maintain controlled progression through turns even when vision and directional memory are unavailable.

\paragraph*{4) Global Dynamics of Turn Negotiation}

\begin{figure}[H]
    \centering
    \includegraphics[width=0.95\linewidth]{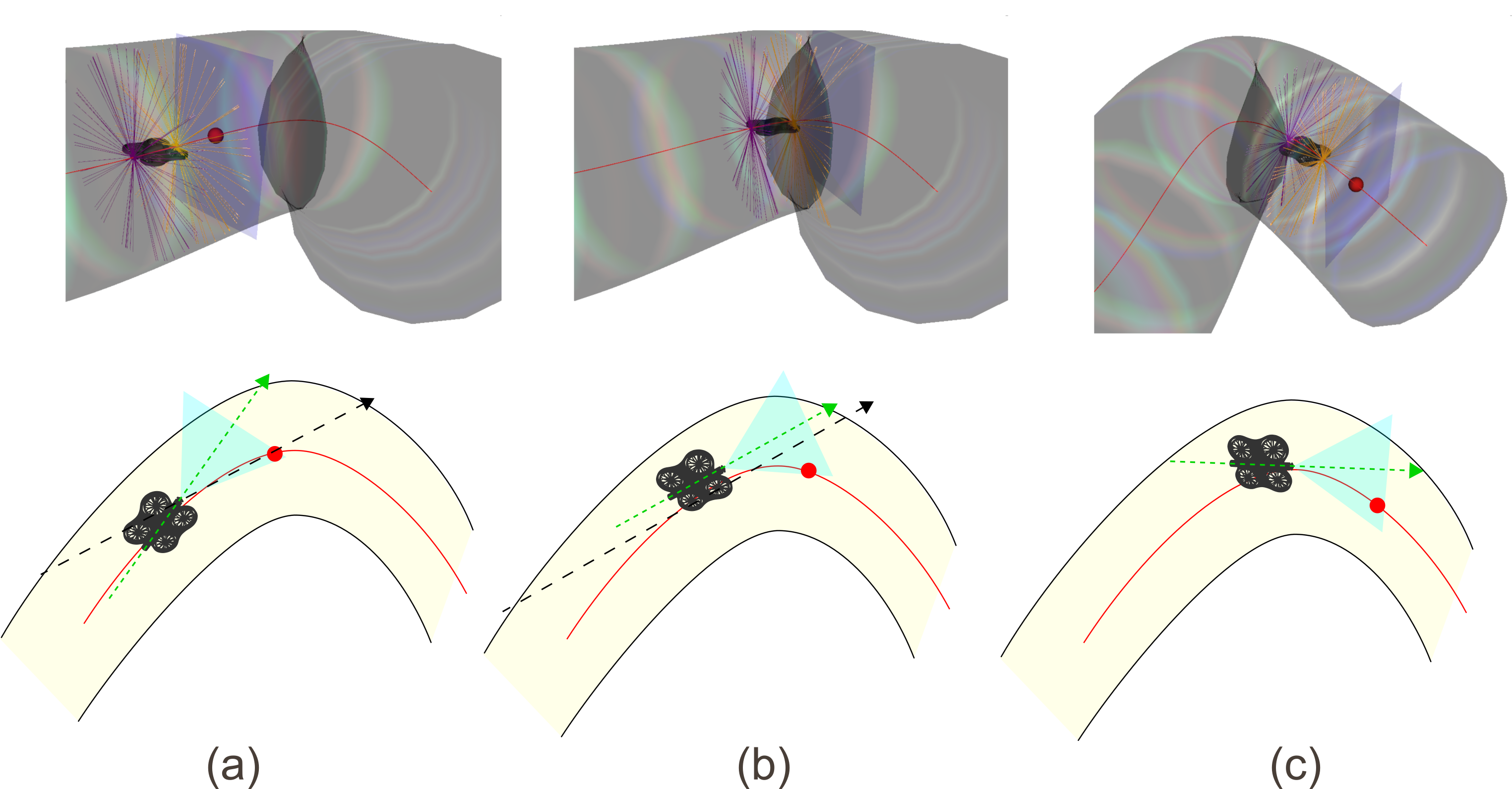}
    \caption{Illustration of the turn negotiation principle.}
    \label{fig:turn_strategy}
\end{figure}

The complete strategy produces a coherent, continuous sequence:
(i) visual anticipation via degradation of frontal alignment,
(ii) directional maintenance using active memory when the target (red dot in Figure~\ref{fig:turn_strategy}) becomes invisible,
(iii) geometric stabilization based on LiDAR measurements when vision is no longer exploitable.

This three-phase dynamic enables smooth, robust, and physically plausible navigation, without requiring explicit knowledge of the local curvature of the conduit.

\begin{enumerate}[label=\textbf{(\alph*)}, leftmargin=1.2cm]
    \item At the entrance of the turn, the target is no longer aligned with the drone’s forward direction (green arrow in Figure~\ref{fig:turn_strategy}). The agent stores the last visible direction of the target (black arrow in Figure~\ref{fig:turn_strategy}) before it exits the camera’s field of view.

    \item When the target temporarily disappears, the agent progresses through the turn by attempting to align with the last recorded visual direction.

    \item Until the target becomes observable again, the drone continues moving forward by relying on LiDAR measurements—especially the rear LiDAR—to prevent collisions with the tube walls.
\end{enumerate}

\subsection*{Transition to Reward Shaping}

The mechanisms described above—visual anticipation of turns, directional maintenance through memory, and geometric stabilization based on front and rear LiDAR—depend on a tight coordination between perception and drone dynamics. For these behaviors to emerge during learning, the reward function must encode each step of the process into exploitable signals.

The next section therefore details the construction of the instantaneous reward components. These explicitly encode:
(i) early turn detection via degradation of visual alignment,
(ii) transition to the memory regime when the target disappears,
(iii) differentiated use of front and rear LiDAR to ensure safe recentering in curves,
(iv) geometric progression and trajectory stability.

The goal of this reward shaping is to provide the drone with signals that are consistent with the navigation principles established in the previous section, enabling it to autonomously reproduce this robust behavior across all encountered configurations.

\subsection{Reward Shaping}

The reward function was designed to encourage robust navigation in a constrained tubular environment, where the 
agent operates with predominantly local and partial perception. It relies on a set of signals derived from the 
LiDAR sensors, the front-facing camera, the memory of the target, and the geometric progression within the tube.  
The instantaneous reward $r_t$ results from a weighted combination of these components, modulated by the local 
context (straight segments, turns, presence or absence of the target in the visual field).

% ============================================================
% LIDAR -------------------------------------------------------
% ============================================================

\subsubsection{LiDAR Features and Turn Detection}

At each step, two LiDARs perpendicular to the drone’s axis (front and rear) provide a set of normalized distances 
in $[0,1]$. These measurements are grouped into four sectors (left, right, top, bottom) for both front and rear 
scans, enabling the definition of horizontal and vertical asymmetries:
\begin{equation}
    \Delta_{\mathrm{FH}} = r_{\mathrm{FR}} - r_{\mathrm{FL}}, \quad
    \Delta_{\mathrm{FV}} = r_{\mathrm{FT}} - r_{\mathrm{FB}},
\end{equation}
\begin{equation}
    \Delta_{\mathrm{RH}} = r_{\mathrm{RR}} - r_{\mathrm{RL}}, \quad
    \Delta_{\mathrm{RV}} = r_{\mathrm{RT}} - r_{\mathrm{RB}},
\end{equation}
where $F$ and $R$ denote \textit{front} and \textit{rear} sectors, and $L$, $R$, $T$, $B$ the \textit{left}, \textit{right}, 
\textit{top}, and \textit{bottom} regions. These asymmetries are used to construct symmetry scores:
\begin{equation}
    S_F = 1 - \tfrac{1}{2} (|\Delta_{\mathrm{FH}}| + |\Delta_{\mathrm{FV}}|), \qquad
    S_R = 1 - \tfrac{1}{2} (|\Delta_{\mathrm{RH}}| + |\Delta_{\mathrm{RV}}|),
\end{equation}
where $S_F, S_R \in [0,1]$ measure the drone’s local centering within the tube cross-section.

Turn presence is estimated through a confidence indicator:
\begin{equation}
    C_{\mathrm{turn}} = \mathrm{clip}\!\left(
        (S_R - S_F) + \max(\mu_R - \mu_F, 0), \; 0, \; 1
    \right),
\end{equation}
where $\mu_F$ and $\mu_R$ are the front/rear average LiDAR distances.  
Thus, $C_{\mathrm{turn}} \approx 0$ in straight segments and $C_{\mathrm{turn}} \approx 1$ in tight turns.

% ============================================================
% METRICS -----------------------------------------------------
% ============================================================

\subsubsection{Centering and Alignment}

Centering is evaluated via the radial distance $d_\perp$ to the tube’s local axis. The corresponding score is:
\begin{equation}
    S_{\mathrm{center}} = \mathrm{clip}\!\left(1 - \frac{d_\perp}{R},\, 0, 1\right),
\end{equation}
where $R$ is the tube radius.

Alignment depends on target visibility:
\begin{equation}
    S_{\mathrm{align}} =
    \begin{cases}
        \mathbf{f}_t \cdot \widehat{\mathbf{d}}_{\mathrm{target}}, & \text{target visible}, \\[4pt]
        \mathbf{f}_t \cdot \widehat{\mathbf{d}}_{\mathrm{mem}}, & \text{target absent but memory available}, \\[4pt]
        \mathbf{f}_t \cdot \mathbf{a}_t, & \text{otherwise},
    \end{cases}
\end{equation}
where $\mathbf{f}_t$ is the drone direction, $\mathbf{a}_t$ the local tube axis, and 
$\widehat{\mathbf{d}}_{\mathrm{target}}$, $\widehat{\mathbf{d}}_{\mathrm{mem}}$ the normalized directions to the visible 
or memorized target. An instantaneous trajectory score is defined as:
\begin{equation}
    S_{\mathrm{traj}} = \tfrac{1}{2} S_{\mathrm{center}} + \tfrac{1}{2} S_{\mathrm{align}}.
\end{equation}

% ============================================================
% ADAPTIVE WEIGHTS -------------------------------------------
% ============================================================

\subsubsection{Adaptive Weighting Depending on Context}

The contributions of front and rear LiDARs are modulated according to turn intensity:
\begin{equation}
    w_F = 0.6 \big(1 - 0.4\,C_{\mathrm{turn}}\big), \qquad
    w_R = 0.4 \big(1 + 0.4\,C_{\mathrm{turn}}\big).
\end{equation}
Thus, in straight segments ($C_{\mathrm{turn}} \rightarrow 0$), front geometry dominates, whereas in turns 
($C_{\mathrm{turn}} \rightarrow 1$), rear measurements become more informative.

% ============================================================
% THREE REGIMES -----------------------------------------------
% ============================================================

\subsubsection{Three Instantaneous Reward Regimes}

Depending on target visibility, the instantaneous reward follows three distinct formulations:

\paragraph*{Case 1 --- target visible}
\begin{equation}
    r_t = 
    S_{\mathrm{align}} 
    + w_F S_F + w_R S_R
    + 0.4\,\mu_F
    + 0.3\,(1 - |\Delta_{\mathrm{FH}}|).
\end{equation}

\paragraph*{Case 2 --- target absent but direction memorized}
\begin{equation}
    r_t =
    S_{\mathrm{align}}
    + w_F S_F + w_R S_R
    + 0.4\,\bigl(1 - |\Delta_{\mathrm{FH}}|\bigr)
    + 0.3\,(1 - |\Delta_{\mathrm{FV}}|).
\end{equation}

\paragraph*{Case 3 --- blind navigation (LiDAR only)}
\begin{equation}
    r_t =
    w_F S_F + w_R S_R
    + 0.4\,\mu_F
    - 0.3\,|\!S_F - S_R\!|.
\end{equation}

A warm-up bonus is added at the beginning of each episode ($t < t_{\mathrm{warmup}}$) to stabilize initial alignment:
\begin{equation}
    r_t \leftarrow r_t +
    0.5\, \bigl(0.7\,S_F + 0.3\,S_R\bigr)\,\bigl(1 - |\!S_F - S_R\!|\bigr)\,(1 - 0.5\,C_{\mathrm{turn}}).
\end{equation}

% ============================================================
% TERMINAL REWARDS --------------------------------------------
% ============================================================

\subsubsection{Terminal Rewards}

Three events terminate an episode:
\[
r_{\mathrm{succ}} = +10, \qquad
r_{\mathrm{fail}} = -10, \qquad
r_{\mathrm{timeout}} = -1.
\]

\section{Experimental Results}

This section presents the full set of results obtained during the development, training, and evaluation of the RL agent. 
The analyses include: (i) training parameters, (ii) performance evolution, (iii) comparison with a reference algorithm, 
and (iv) validation in a simulated 3D environment.

%-------------------------------------------------------
\subsection{Training Parameters}

\subsubsection{Model Architecture}

The model uses a \textsc{PPO} architecture composed of a fully connected network that receives as input an 84-dimensional 
observation vector, including simulated LiDAR distances, relative position inside the tube, and drone velocity.  
The architecture consists of:
\begin{itemize}
    \item two dense layers with 256 and 128 neurons, respectively;
    \item \texttt{tanh} activation for the main layers and \texttt{relu} for the final layers;
    \item observation normalization;
    \item an output layer producing continuous actions (linear and angular velocities).
\end{itemize}

\subsubsection{Hyperparameters}

Training was performed using the \textsc{Ray} library (version~2.49.2) and its \textsc{RLlib} implementation of the PPO paradigm.  
The main training hyperparameters are summarized in Table~\ref{tab:hparams}. These parameters result from an initial 
exploration aimed at ensuring return-signal stability and sufficient success rates.

\begin{table}[h!]
\centering
\caption{Training hyperparameters for the PPO model.}
\label{tab:hparams}
\begin{tabular}{l c}
\hline
\textbf{Parameter} & \textbf{Value} \\
\hline
Algorithm & PPO \\
Learning rate & $3\times 10^{-4}$ \\
Gamma & $0.99$ \\
Train batch size & $13000$ \\
Mini-batch size & $1300$ \\
Rollout fragment length & $200$ \\
Number of actors (env runners) & 13 \\
Number of environments per actor & 1 \\
Entropy coefficient & 0.003 \\
Clip parameter & 0.3 \\
\hline
\end{tabular}
\end{table}

\subsubsection{Computation Resources in Python Environment}

Training was conducted on a server machine equipped with:

\begin{itemize}
    \item \textbf{CPU}: 2 × AMD EPYC 7F52 (32 physical cores each);
    \item \textbf{GPU}: 4× NVIDIA (driver 535.230.02), 2 of which were used;
    \item \textbf{OS}: Ubuntu 20.04.6 LTS;
    \item \textbf{Python version}: 3.9.5.
\end{itemize}

%-------------------------------------------------------
\subsection{Training Evaluation}

\begin{figure}[h!]
\centering
\includegraphics[width=0.9\linewidth]{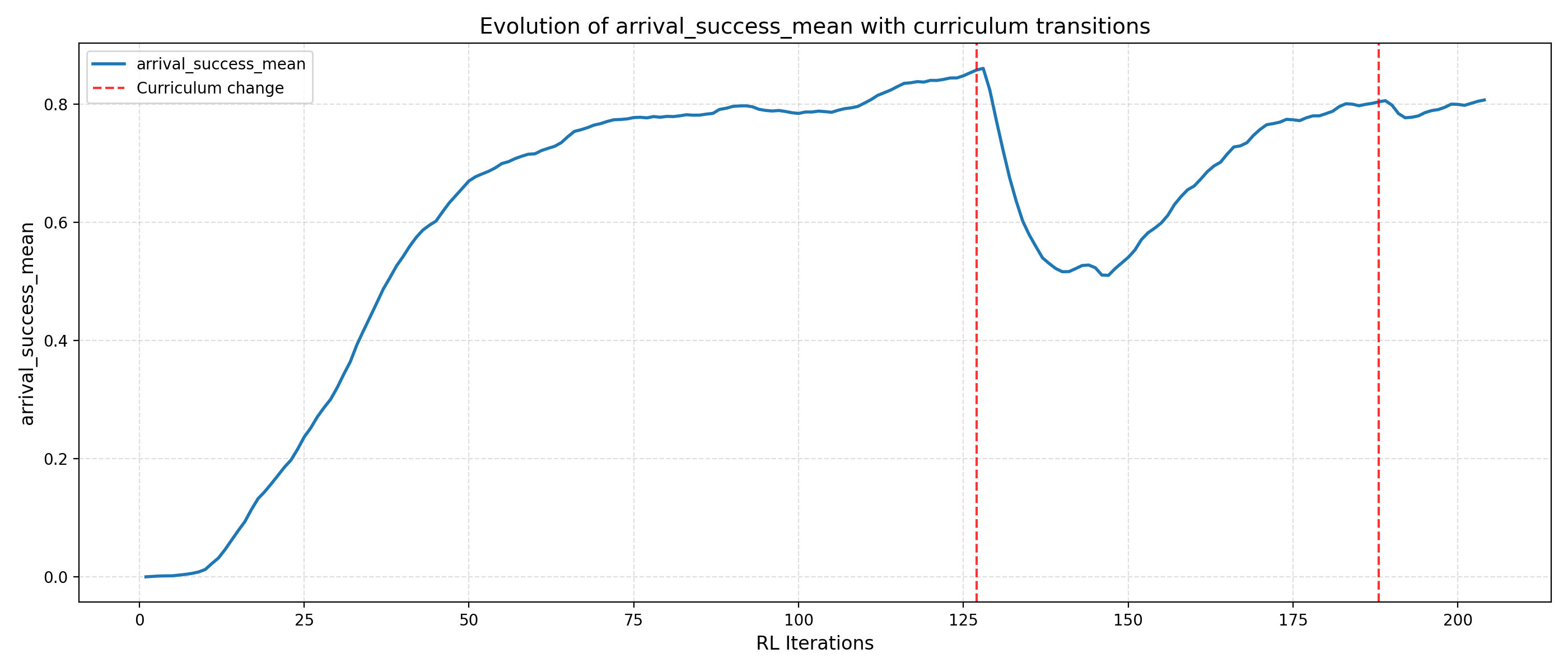}
\caption{Average success during training (arrival success).}
\label{fig:success}
\end{figure}

\begin{figure}[h!]
\centering
\includegraphics[width=0.9\linewidth]{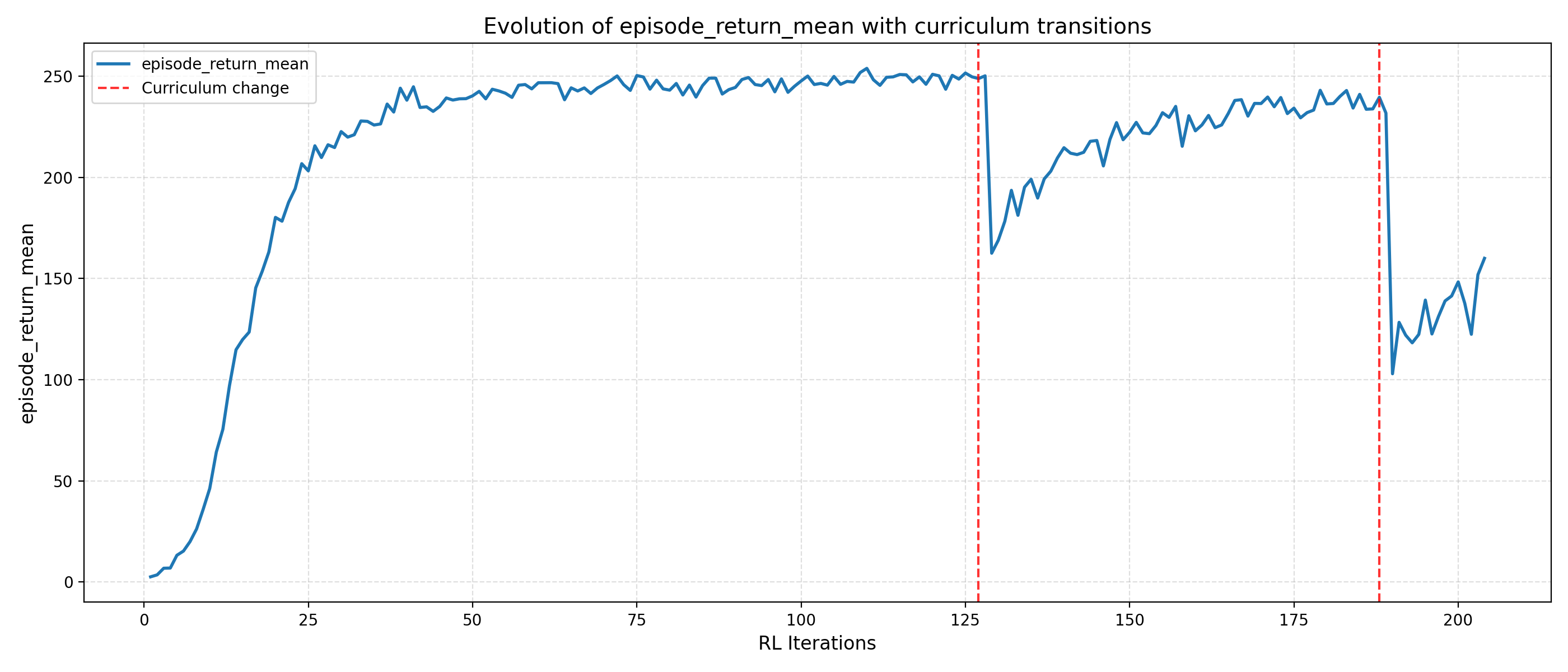}
\caption{Average return during training.}
\label{fig:return}
\end{figure}

The agent’s performance is evaluated using two metrics:  
(i) the \textit{average success rate}, defined as the proportion of episodes in which the drone reaches the end of the tube without collision;  
(ii) the \textit{average return}, corresponding to the cumulative sum of rewards obtained in each episode.  
The evolution of these indicators is shown in Figures~\ref{fig:success} and~\ref{fig:return}.

\subsubsection{Average Success per Iteration}

Figure~\ref{fig:success} illustrates the progression of the success rate throughout training. A rapid performance increase is observed during the first few dozen iterations, corresponding to the learning of the simplest curriculum scenarios (near–rectilinear tubes).

After reaching the second curriculum level, a noticeable drop in success rate appears, reflecting the increased difficulty caused by tighter turns and temporary loss of the vanishing point. This initial degradation is followed by a gradual recovery, indicating that the agent progressively adapts to these more demanding conditions, eventually reaching stabilized behavior before advancing to the final level.

On the last level of the curriculum, the agent stabilizes around an average success rate between \textbf{0.75} and \textbf{0.80}, indicating robust convergence in the most complex environments. This stabilization confirms the ability of the learned policy to handle strongly curved tube geometries, where navigation requires anticipating the trajectory from partial or intermittent visual cues.

\subsubsection{Average Return per Episode}

Figure~\ref{fig:return} shows the evolution of the average return. The curve exhibits steady growth during the early curriculum levels, followed by stabilization phases as complexity increases.

Localized variations in return, particularly noticeable during level transitions, reflect the increasing difficulty of the environments: the tighter the turns, the more the policy must finely adjust its direction while avoiding collisions. Nevertheless, variance gradually decreases over the course of training, indicating an increasingly consistent and reproducible policy.

The final stabilization of the return confirms that the policy adapts well to the most demanding environments in the curriculum, highlighting the effectiveness of the proposed reward shaping.

\medskip
It is worth noting that the average return decreases in the final curriculum level, stabilizing around values close to 150, whereas it remained between 200 and 250 in earlier levels.
This reduction does not indicate a degradation of the learned policy, but rather reflects the intrinsic difficulty of highly curved tube geometries: episodes tend to be shorter, corrective maneuvers are more frequent, and the reward accumulated per time step becomes structurally lower even when the agent succeeds.

Moreover, due to the moderate success threshold used to trigger progression between levels, the agent performs only a limited number of iterations on the final level. As a consequence, the policy receives fewer optimization steps in the most demanding configuration, which partly explains why the return does not rise to the same values as in simpler levels, despite a comparable success rate.

These effects are therefore expected consequences of the reward structure and curriculum design rather than signs of instability in the learning process.

\subsubsection*{Impact of Curriculum Learning}

The presented performances must be interpreted in light of the \emph{Curriculum Learning} strategy employed.  
In this first study, the objective is not to maximize the absolute success rate but to ensure feasibility, stability, and convergence of learning across tubular environments of increasing complexity. For this purpose, the success thresholds governing progression between curvature levels (Table~\ref{tab:curriculum_tubes}) were deliberately set to moderate values.

A stricter choice of thresholds (e.g., $>0.9$) would have increased the risk of stagnation in intermediate levels, especially when curvature leads to intermittent loss of the vanishing point. Excessively high success requirements at this stage would have lengthened training disproportionately, without providing substantial benefit for this proof of concept.

Conversely, the chosen thresholds allow for regular progression while giving the agent sufficient time to adapt to the specific characteristics of each difficulty level. The fluctuations observed in the performance curves therefore do not reflect instability in training, but rather the expected transitions between distinct navigation regimes.

\medskip
\noindent Overall, the results show that:
\begin{enumerate}
\item the agent quickly acquires basic skills in simple configurations,
\item it succeeds in adapting to levels where visual information becomes intermittent or partial,
\item the policy converges in the most demanding environments of the curriculum.
\end{enumerate}

This behavior validates the relevance of combining a directional action space, reward shaping, and controlled difficulty progression, as well as the suitability of the chosen thresholds for this initial demonstration.

%-------------------------------------------------------
\subsection{Comparison with the Reference Algorithm}

In this section, we experimentally evaluate the performance of the proposed PPO agent by comparing it with the deterministic \textsc{Pure Pursuit} (PP) algorithm, commonly used for trajectory tracking in mobile robotics. The comparison is carried out over the three complexity levels defined in the curriculum, each consisting of 100 independent episodes.

\subsubsection*{Comparison Framework and Information Asymmetry}

It is essential to emphasize that the two methods do not operate under the same informational conditions.  
Pure Pursuit benefits from a major structural advantage: the algorithm has \emph{a priori} access to the exact geometric axis of the tube, which it uses as a reference to compute a tracking command. The ideal trajectory is therefore provided explicitly and with perfect accuracy.

In contrast, the PPO agent has no information about the true geometry of the tube. It must infer the direction of progression from partial observations:
\begin{itemize}
    \item local LiDAR measurements describing only wall proximity;
    \item visual detection of the tube center when it is within the camera’s field of view;
    \item implicit estimation of trajectory continuity when the vanishing point disappears in strongly curved regions.
\end{itemize}

Thus, PP operates in a context of \emph{complete information}, whereas PPO must solve a navigation problem in a partially observable environment. The fact that PPO exceeds PP on several metrics should therefore be interpreted in light of this fundamental asymmetry.

\subsubsection*{Comparison Results}

The results obtained are presented in Table~\ref{tab:comparison_real}. The selected metrics are:  
(i) success rate,  
(ii) tube-exit rate (failure without collision but loss of confinement),  
(iii) a navigation quality index aggregating centering and alignment.

\begin{table}[h!]
\centering
\caption{Experimental comparison between PPO and Pure Pursuit over three complexity levels (100 episodes per level).}
\label{tab:comparison_real}
\begin{tabular}{c|ccc|ccc}
\hline
 & \multicolumn{3}{c|}{\textbf{Pure Pursuit}} & \multicolumn{3}{c}{\textbf{PPO (proposed)}} \\
\textbf{Level} &
Success (\%) & Exits (\%) & Quality &
Success (\%) & Exits (\%) & Quality \\
\hline
0 & 66 & 34 & 0.699 & \textbf{84} & \textbf{16} & 0.609 \\
1 & 34 & 66 & 0.553 & \textbf{71} & \textbf{29} & 0.526 \\
2 & 73 & 27 & 0.532 & \textbf{87} & \textbf{13} & 0.522 \\
\hline
\end{tabular}
\end{table}

\paragraph{Definition of the Navigation Quality Index.}

The navigation quality index $Q$ used in our study is directly derived from the measurements collected at each simulation step. At every instant, two metrics are computed when the drone is inside the tube:

\begin{itemize}
    \item a \textbf{centering} score $d_c \in [0,1]$, defined as
    \[
    d_c = 1 - \frac{r_\perp}{R},
    \]
    where $r_\perp$ is the radial distance of the drone to the local tube axis and $R$ the tube radius;
    \item an \textbf{alignment} score $a \in [-1,1]$, corresponding to the dot product between the drone’s direction and the estimated tube or vanishing-point direction (via camera or memory), normalized to $[0,1]$.
\end{itemize}

At the end of the episode, average values $\overline{d}_c$ and $\overline{a}$ are computed, and the quality index is defined as an equally weighted average:
\[
Q = \frac{1}{2}\,\overline{d}_c
  + \frac{1}{2}\,\overline{a},
\qquad Q \in [0,1].
\]

A high value of $Q$ corresponds to a trajectory that is well centered and well aligned with the local tube direction, regardless of whether the episode ends in success or failure. It should be noted that this index is not a global performance measure, but rather an indicator of the \emph{geometric cleanliness} of the trajectory.

\subsubsection*{Performance Analysis}

\paragraph{Success Rate.}

At all curriculum levels, PPO achieves a substantially higher success rate than PP.  
The gap is especially pronounced at the intermediate level (71\% vs. 34\%), where tube curvature strongly disrupts Pure Pursuit due to its reliance on continuous visibility of the ideal trajectory.

\paragraph{Robustness and Tube Exits.}

PPO consistently and significantly reduces the tube-exit rate:
\begin{itemize}
    \item 34\% $\rightarrow$ 16\% at level 0,
    \item 66\% $\rightarrow$ 29\% at level 1,
    \item 27\% $\rightarrow$ 13\% at level 2.
\end{itemize}
This indicates that the learned policy is better equipped to correct trajectories when the environment imposes tight turns or when the tube center is not visible.

\paragraph{Interpretation of the Quality Index.}

While Pure Pursuit achieves a slightly higher average quality index—reflecting more aligned and centered paths—this difference remains moderate and must be interpreted cautiously.  
PP benefits from direct access to the ideal trajectory, allowing it to produce cleaner movements, but this does not guarantee its ability to remain within the tube when geometry becomes complex.

In contrast, PPO may produce less smooth trajectories at times, but systematically prioritizes viable navigation. This strategy results in a far higher success rate at all levels: the RL agent maintains the drone inside the tube even when visual cues become ambiguous or when curvature forces PP into maneuvers it cannot anticipate.

\subsubsection*{Summary and Significance of the Results}

In a context where only the PPO agent must implicitly infer the tube structure from sparse sensor signals, while Pure Pursuit relies on complete geometric knowledge, the results are significant:

\begin{enumerate}
    \item PPO exceeds Pure Pursuit in success rate across \emph{all} complexity levels;  
    \item PPO exhibits more robust navigation, with a clear reduction in tube exits;  
    \item the learned agent demonstrates an ability to implicitly reconstruct the direction of progression, even in zones where visual information is partial or absent.  
\end{enumerate}

These findings show that the reinforcement learning approach does more than merely imitate Pure Pursuit: it acquires a navigation capability that is \emph{not accessible} to a geometric controller relying on full model knowledge.  
This opens promising perspectives for autonomous navigation in unknown or unmodeled tubular environments, such as natural tunnels, industrial infrastructures, or anatomical channels.

%-------------------------------------------------------
\subsection{Validation in a High-Fidelity Simulated Environment}

To test the agent in a setting close to real-world conditions, we developed a Unity scene designed to provide a faithful three-dimensional visualization of the agent’s navigation inside an industrial-type tubular conduit. Although perception and interactions with the tube walls are computed on the Python side to remain consistent with the training environment, the Unity 3D engine provides a physically coherent simulation based on a rigid-body model subject to gravity. This integration makes it possible to obtain and visualize a continuous inertial evolution of the drone under the control commands issued by the agent’s policy.

At this stage, the visualization serves primarily as a complementary tool to qualitatively illustrate the learned behavior.  
Future steps will include:
(i) integrating a more realistic sensor model (camera, measurement noise),  
(ii) increasing the fidelity of the drone dynamics in Unity,  
(iii) conducting tests in real geometries digitized or reconstructed from 3D scans.

\subsubsection{Description of the Unity Physical Model}

The drone’s motion is described using classical Newtonian equations governing the translation and rotation of a rigid body. Linear dynamics are modeled as:
\begin{equation}
m \, \dot{\mathbf{v}} = \mathbf{F}_{\mathrm{cmd}} + m \, \mathbf{g},
\end{equation}
where $m$ is the drone mass, $\mathbf{v}$ its linear velocity, $\mathbf{g}$ the gravity vector, and $\mathbf{F}_{\mathrm{cmd}}$ the force resulting from the RL command transformed by Unity (desired direction and speed level).

Orientation is updated using rigid-body rotational dynamics:
\begin{equation}
\mathbf{I} \, \dot{\boldsymbol{\omega}} = \boldsymbol{\tau}_{\mathrm{cmd}}
- \boldsymbol{\omega} \times (\mathbf{I} \boldsymbol{\omega}),
\end{equation}
where $\boldsymbol{\omega}$ is the angular velocity, $\mathbf{I}$ the inertia tensor, and $\boldsymbol{\tau}_{\mathrm{cmd}}$ the torque derived from the agent’s directional command.

Unity thus acts as a robust inertial integration engine: at each control interval, it computes the next pose $\mathbf{x}_{t+\Delta t}$ from $\mathbf{x}_t$ and the applied command, ensuring dynamic continuity, gravity effects, and numerically stable orientation updates.

\paragraph*{Data Exchange Between Python and Unity.}

The actions produced by the agent specify only a desired movement direction in the drone’s visual space and a speed level. Before being transmitted to the Unity physics engine, this direction is normalized and may be adjusted by a small stabilizing vertical term. This term’s only purpose is to partially counteract gravity, preventing a drone initially in hover from immediately dropping in the absence of explicit thrust commands. This preprocessing step does not modify the underlying dynamic model: it merely translates the agent’s abstract action into the force $\mathbf{F}_{\mathrm{cmd}}$ applied to the rigid body.

The physical simulation itself is fully managed by Unity. At each control step requested by Python, Unity performs exactly one dynamic integration according to Newtonian equations, simultaneously applying the command force, gravity, inertial effects, and any potential contacts with the environment. Gravity is therefore intrinsic to Unity’s physics engine and is applied automatically at every simulation step, independent of the control frequency imposed by Python.

After this integration step, Unity sends back to Python the resulting dynamic state, consisting of the drone’s 3D position, its orientation as a quaternion, and its linear and angular velocities. This information forms the physical basis from which Python reconstructs the drone’s local reference frame and feeds all perception and evaluation modules.

Meanwhile, perception, tube geometry, and progression evaluation remain entirely handled on the Python side. This environment is responsible for:
(i) computing the derived LiDAR measurements,  
(ii) verifying that the drone remains inside the tubular structure,  
(iii) estimating longitudinal progression,  
(iv) detecting episode termination conditions.

\subsubsection{Exchange Protocol Between Python and Unity via Shared Memory}

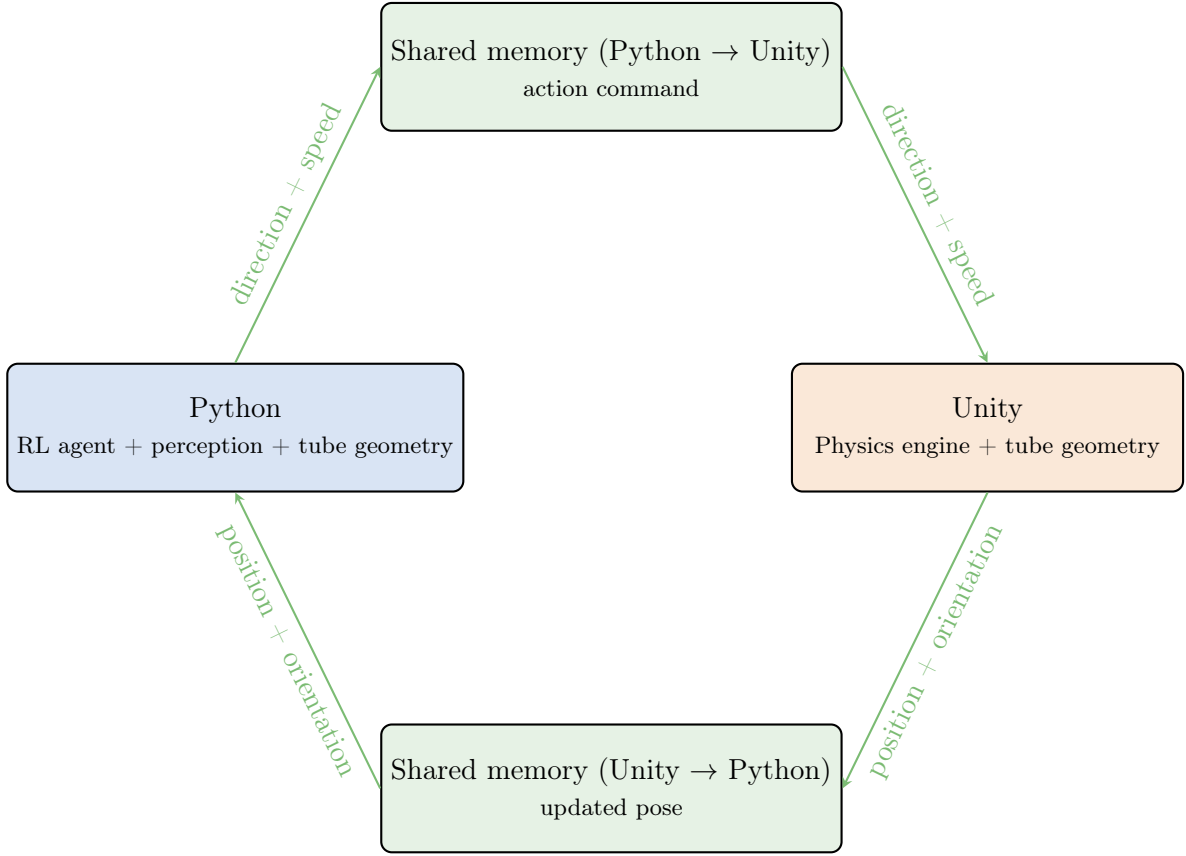
\begin{figure}[h!]
\centering
\begin{tikzpicture}[>=stealth, thick]

% Colors
\definecolor{pythonblue}{RGB}{66,120,200}
\definecolor{unityorange}{RGB}{230,140,60}
\definecolor{memgreen}{RGB}{90,170,80}

% Node positions (cross layout, medium size)
\node[draw, rounded corners, fill=pythonblue!20,
      minimum width=5.2cm, minimum height=1.7cm,
      align=center, thick]
(py) at (-5,0) {Python\\\footnotesize RL agent + perception + tube geometry};

\node[draw, rounded corners, fill=unityorange!20,
      minimum width=5.2cm, minimum height=1.7cm,
      align=center, thick]
(unity) at (5,0) {Unity\\\footnotesize Physics engine + tube geometry};

\node[draw, rounded corners, fill=memgreen!15,
      minimum width=5.2cm, minimum height=1.7cm,
      align=center, thick]
(memPYU) at (0,4.8)
{Shared memory (Python $\rightarrow$ Unity)\\\footnotesize action command};

\node[draw, rounded corners, fill=memgreen!15,
      minimum width=5.2cm, minimum height=1.7cm,
      align=center, thick]
(memUPY) at (0,-4.8)
{Shared memory (Unity $\rightarrow$ Python)\\\footnotesize updated pose};

% --- Arrows with sloped labels ---

\draw[->, thick, memgreen!80]
  (py.north) -- (memPYU.west)
  node[midway, above, sloped, align=center]
    {direction + speed};

\draw[->, thick, memgreen!80]
  (memPYU.east) -- (unity.north)
  node[midway, above, sloped, align=center]
    {direction + speed};

\draw[->, thick, memgreen!80]
  (unity.south) -- (memUPY.east)
  node[midway, below, sloped, align=center]
    {position + orientation};

\draw[->, thick, memgreen!80]
  (memUPY.west) -- (py.south)
  node[midway, below, sloped, align=center]
    {position + orientation};

\end{tikzpicture}
\caption{Bidirectional communication protocol between Python and Unity via two shared memory buffers.}
\label{fig:memory_exchange}
\end{figure}

As illustrated in Figure~\ref{fig:memory_exchange}, real-time communication between the Python environment (which computes the action and navigation indicators) and Unity (which integrates inertial dynamics) relies on a shared-memory mechanism.  
This approach enables bidirectional, very low-latency communication without network overhead, ensuring precise synchronization between the two execution engines.  
It further benefits from relying solely on native functionalities: \texttt{MemoryMappedFile} in C\# for Unity and \texttt{multiprocessing.shared\_memory} in Python, avoiding external dependencies and ensuring maximal portability.

Two shared-memory buffers are used:

\begin{itemize}
    \item \textbf{a Python $\rightarrow$ Unity buffer} containing the instantaneous command issued by the RL policy, already adjusted to the Unity physical environment (desired direction and forward speed);
    \item \textbf{a Unity $\rightarrow$ Python buffer} storing the updated drone pose produced by the physics engine (position, orientation, and optionally velocities).
\end{itemize}

At each control step, Python writes into the first buffer the action computed by the agent, which Unity can immediately read and interpret as a physical command applied to the drone’s rigid body.  
Symmetrically, Unity writes the updated pose into the second buffer after performing inertial integration over the interval $\Delta t$.  
Python then reads this pose in order to:

(i) estimate the drone’s progression along the tube’s centerline;  
(ii) evaluate centering and alignment metrics;  
(iii) verify trajectory validity (remaining inside the tube, no collision);  
(iv) construct the next observation fed to the agent.

This shared-memory protocol offers several advantages:  
(1) it avoids any network overhead,  
(2) it guarantees constant and minimal access latency,  
(3) it enables a high control frequency compatible with a real-time physics engine,  
(4) it relies exclusively on standard mechanisms already available in both environments.

The Python–Unity coupling thus operates as a synchronous closed loop: Python writes the command, Unity executes the dynamics and writes the updated pose, and Python interprets this pose to produce the next action.

\subsubsection{Use of the Blender 3D Modeling Tool}

\begin{figure}[h!]
    \centering
    \includegraphics[width=1.0\linewidth]{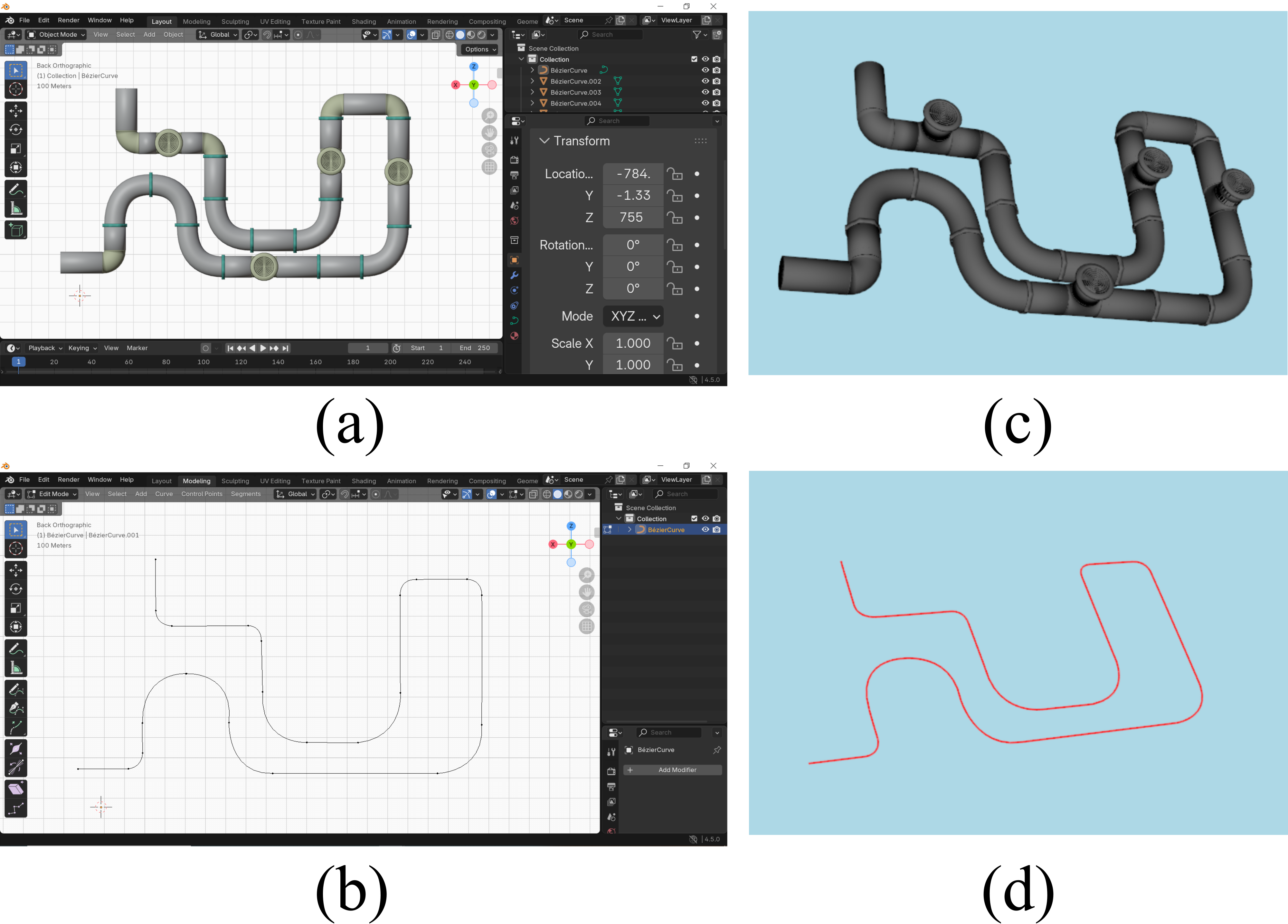}
    \caption{(a) View of the industrial tubular structure in Blender - (b) View of the industrial tubular structure in Python window (Vedo package) - (c) View of the industrial tubular structure's geometric centerline in Blender - (d) View of the industrial tubular structure's geometric centerline in Python window (Vedo package).}
    \label{fig:blender}
\end{figure}

As illustrated in Figure~\ref{fig:blender}, we modeled the industrial tubular conduit and its geometric centerline in Blender, then exported them separately in \texttt{.obj} format.  
The two environments make use of these files in the following way:

\begin{itemize}
    \item \textbf{In Unity}, the \texttt{.obj} file of the tubular conduit is imported directly to provide a realistic 3D visualization of the drone’s inertial dynamics inside the structure;
    \item \textbf{In Python}, both the \texttt{.obj} conduit file and its centerline are loaded to manage tube-center perception and wall interactions during inference.
\end{itemize}

\newpage

\subsubsection{Inference Algorithm}

\begin{figure}[h!]
    \centering
    \includegraphics[width=1.0\linewidth]{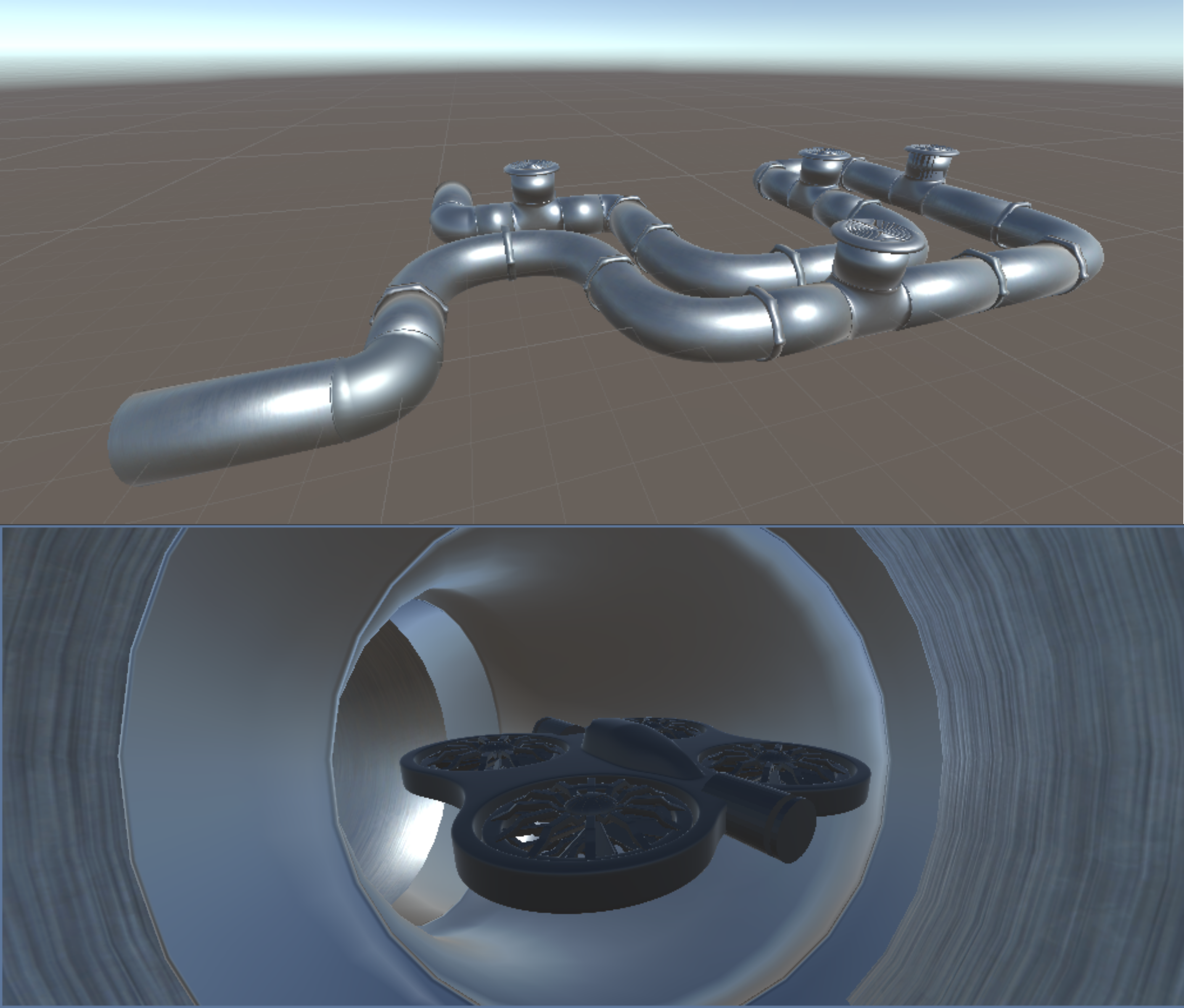}
    \caption{Illustration of the drone’s motion inside an industrial tubular structure in the Unity scene during inference.}
    \label{fig:unity_visualization}
\end{figure}

The complete inference process of the drone with Unity-assisted inertial integration is summarized in the following algorithm.  
Figure~\ref{fig:unity_visualization} shows the motion of the drone inside the industrial tubular structure within the Unity 3D scene.

\bigskip

\begin{algorithm}[h]
\caption{Inference of the trained navigation agent with Unity-assisted physical integration}

Set the control time step $\Delta t$ and Unity’s internal simulation time step $\delta t$\;

Load in Python the trained navigation policy\;

Load in Python the tubular mesh (\texttt{.obj}) and the geometric centerline (\texttt{.obj})\;

Send the initial drone pose to Unity with gravity-compensation correction\;

\For{timestep $= 1, 2, \ldots$}{    

    Compute the current observation from Python  
    (LIDAR-derived features, orientation metrics,  
    tube-relative coordinates, visibility cues)\;

    Generate the action prescribed by the trained policy  
    (desired forward direction and forward speed)\;

    Send the gravity-compensated action command to Unity\;

    \For{$i = 1$ \KwTo $\Delta t / \delta t$}{
        Simulate the drone motion in Unity’s physical engine  
        (inertial update, gravity, orientation adjustment)\;
    }

    Obtain the updated drone position and orientation from Unity\;

    Evaluate the geometric interaction with the tube in Python  
    (deviation from centerline, boundary proximity, forward progression)\;

    \If{exit, collision, or time-limit condition occurs}{
        terminate the episode\;
    }

}
\end{algorithm}

\FloatBarrier
\subsection{Summary}

The experimental results demonstrate that the reinforcement learning approach adopted in this work enables reliable autonomous navigation in tubular environments of increasing complexity.  
The agent progressively acquires an effective strategy thanks to Curriculum Learning, which facilitates the transition from simple rectilinear scenarios to highly curved conduits where the tube center becomes intermittently visible.  
The observed convergence, between 75\% and 80\% success in the most demanding environments, shows the policy’s ability to anticipate the local geometry from partial visual and LiDAR cues.

The comparison with the deterministic \textsc{Pure Pursuit} algorithm highlights a central point: despite a major information asymmetry—PP has exact knowledge of the tube’s geometric axis while PPO must infer it indirectly—the learned policy systematically outperforms the classical controller in success rate and robustness.  
The trajectories produced by PPO are sometimes less smooth, but they allow the drone to remain confined within the tube in situations where geometry or visibility make strict tracking of an ideal trajectory ineffective.

Finally, inertial validation in Unity confirms that the learned strategy remains functional in a more realistic physical setting than that used during training.  
The agent maintains stable trajectories under continuous dynamics and gravity, showing that the learned behaviors do not rely on specific kinematic simplifications.

Overall, the results validate the combined relevance of  
(i) a directionally constrained yet expressive action space,  
(ii) reward shaping tailored to confined environments, and  
(iii) a controlled increase in task complexity.  
The trained agent proves not only effective but also capable of generalizing to situations in which a controller relying on explicit geometric modeling fails.

\section{Discussion and Future Perspectives}

The results obtained highlight several design choices that contributed to the success of the approach, while also revealing limitations that may guide future research.

\subsection{Action Space and Modeling Limitations}

The discrete action space used in this work was specifically designed for navigation in narrow tubular environments.  
By restricting orientations to those lying within the drone’s visual cone and pointing forward, it avoids dangerous or irrelevant maneuvers while reducing learning complexity.  
This pragmatic choice enabled the rapid emergence of stable behaviors.

However, this action space also presents limitations.  
First, the direct coupling between perception and action—only visually plausible directions are allowed—simplifies the problem but does not reflect the capabilities of a real quadrotor, which can initiate maneuvers outside its field of view.  
Second, the drone dynamics were modeled in an essentially kinematic manner during training.

These simplifications provide a solid foundation but naturally call for extensions: densification or regularization of the directional grid, introduction of continuous actions, or integration of a richer dynamic model allowing explicit handling of forces and torques.

Importantly, the proposed framework does not account for aerodynamic disturbances induced by close proximity to walls, such as lateral ground effects, flow deviations, or turbulence generated in narrow sections.  
These phenomena, common in confined environments, can significantly affect the real stability of a quadrotor and are absent from the current model.  
Incorporating them—whether via simplified aerodynamic simulators or experimental data—represents a key perspective for bringing the agent closer to real operational conditions.

\subsection{Turn Negotiation and Partial Information Handling}

Turn negotiation is the most challenging scenario in a tubular environment.  
The current strategy relies on a combination of three mechanisms:  
(i) direct use of the visual target when available,  
(ii) a memory mode that temporarily prolongs the previous direction,  
(iii) exploitation of lidar symmetries to detect and follow local geometry.

This approach has proven effective, but it also shows limitations.  
The use of directional memory may lead to suboptimal behaviors when geometry changes abruptly, and lidar symmetry measures are sensitive to complex or anisotropic structures.  
Natural extensions include the introduction of recurrent models (LSTM or GRU), explicit estimation of the tube’s local curvature, or richer 3D perception to better characterize the surrounding geometry.

\subsection{Visual Perception: From Ideal Signal to Real Onboard Vision}

In this work, the direction of the tube center is assumed to be detected ideally whenever it appears in the field of view.  
This assumption facilitates analysis of the navigation behavior but does not reflect the challenges of real onboard imaging.  
Illumination, textures, reflections, or motion blur strongly affect the visual estimation of a vanishing point or main conduit direction.

A key perspective is therefore to replace this idealized module with a real estimator, based for example on segmentation networks, geometric detection pipelines, or an end-to-end perception–control approach.  
Such a transition would bring the system closer to real operational conditions while enabling co-adaptation between learned perception and navigation strategy.

\subsection{Generalization and Extension to Other Tubular Environments}

Although developed for aerial navigation in industrial, natural, or underground environments, the proposed framework has a much broader scope.  
The challenges addressed here—progression in a confined conduit, partial local perception, intermittent loss of visibility, curvature anticipation—also arise in medical applications, particularly in endoscopy or robotic navigation inside anatomical canals.  
In such contexts, the mechanisms developed here (partial information handling, turn negotiation, compact geometric state representation) form a promising basis for miniature or semi-autonomous navigation approaches.

\subsection{Global Perspectives}

Future work may focus on:
\begin{itemize}
    \item integrating realistic real or synthetic visual perception;
    \item exploring continuous and dynamically coherent action spaces;
    \item incorporating full drone dynamics during training;
    \item extending to even more complex tubular geometries (helical or branched);
    \item experimental validation on a real robotic platform.
\end{itemize}

Thus, the methodology presented constitutes a robust initial foundation for learning autonomous behaviors in confined environments, while opening pathways toward transdisciplinary developments ranging from industrial robotics to miniature medical robotics.

\section{Conclusion}

This work presents a reinforcement learning approach for autonomous drone navigation in confined tubular environments, a context characterized by restrictive geometry, partial perception, and intermittent visibility loss.  
The proposed framework—directional action space, tailored reward shaping, Curriculum Learning progression, and inertial integration through Unity—enabled the training of a robust policy capable of generalizing to unknown geometries.

The results show that the PPO agent learns stably to negotiate both rectilinear and highly curved conduits, achieving success rates between 75\% and 80\% in the most complex scenarios.  
The comparison with the \textsc{Pure Pursuit} algorithm reveals a key insight: despite a significant information asymmetry—PP has access to the exact geometric axis of the tube, whereas PPO must infer it solely from lidar and visual cues—the learned policy consistently outperforms the classical controller in robustness and tube retention.  
These results demonstrate that the agent can implicitly reconstruct the direction of progression, even when the conduit center disappears from the field of view in sharp turns.

The 3D validation in Unity confirms that the learned behavior remains effective under continuous inertial dynamics, representing an important step toward deployment in real-world conditions.  
The proposed approach therefore provides a unified framework, spanning from kinematic learning to realistic physical evaluation, while maintaining geometric consistency through the use of a shared 3D model.

Beyond industrial or natural tubular environments, the developed methodology has broader applicability: navigation in confined spaces, automated inspection, subterranean robotics, or even guidance within anatomical channels in medical robotics.  
The mechanisms studied—partial perception management, turn negotiation, local geometry anticipation—are directly transferable to these fields.

Future perspectives include integrating more realistic real or synthetic visual perception, adopting continuous or dynamically coherent action spaces, using recurrent models for improved partial-information handling, testing on more complex or branching tubular geometries, and accounting for aerodynamic disturbances inherent to confined environments.  
Ultimately, experimentation on a real robotic platform will constitute the decisive step in validating the operational applicability of the approach.

Overall, this work shows that a reinforcement learning agent can not only master navigation in an unknown conduit but also surpass a classical solution built upon complete geometric knowledge.  
It thus opens the path toward autonomous systems capable of operating reliably in confined environments where explicit modeling is difficult or impossible.

%\section*{Acknowledgments}

% À compléter : financements, collaborations, remerciements

\section*{Author Contributions}
Main manuscript writing, Z. Mari; design, development, and evaluation of the reinforcement learning agent, Z. Mari; 
review and proofreading, J. Pasquet and J. Seinturier;
project coordination and scientific supervision, Z. Mari.

\newpage
%\section{Références}  

\end{document}